\documentclass[lettersize,journal]{IEEEtran}

\usepackage{amsmath,amsfonts}
\DeclareMathOperator*{\argmax}{arg\,max}

\usepackage{algorithmic}
\usepackage{algorithm}
\usepackage{array}
\usepackage[caption=false,font=normalsize,labelfont=sf,textfont=sf]{subfig}

\usepackage{textcomp}
\usepackage{stfloats}
\usepackage{url}
\usepackage{verbatim}
\usepackage{graphicx}
\usepackage{cite}

\usepackage[utf8]{inputenc} %
\usepackage[T1]{fontenc}    %
\usepackage{CJKutf8}

\usepackage{amssymb}
\usepackage{mathtools}
\usepackage{amsthm}

\usepackage{graphicx}
\usepackage{wrapfig}
\usepackage[caption=false,font=normalsize,labelfont=sf,textfont=sf]{subfig}

\usepackage{booktabs}
\usepackage{multirow}
\usepackage{makecell}

\usepackage{algorithm}
\usepackage{algorithmic}

\usepackage{array}
\usepackage{textcomp}
\usepackage{stfloats}
\usepackage{verbatim}
\usepackage{url}
\usepackage{enumitem}
\usepackage{dashrule}
\usepackage{titletoc}
\usepackage{lipsum}

\usepackage{bbm}
\usepackage{dsfont}

\usepackage{stackengine}

\usepackage[table,xcdraw,svgnames,HTML]{xcolor}
\usepackage{hyperref}
\usepackage[capitalize]{cleveref}

\usepackage{tikz}
\usetikzlibrary{tikzmark}

\usepackage[most]{tcolorbox}
\tcbuselibrary{breakable}

\usepackage{nicematrix}

\providecommand{\citep}{\cite}
\providecommand{\citet}{\cite}

\definecolor{mydarkgreen}{RGB}{0,100,0}
\definecolor{light-gray}{gray}{0.6}
\colorlet{ourred}{red!4}
\colorlet{ourblue}{blue!100}
\definecolor{mediumpersianblue}{rgb}{0.0, 0.4, 0.65}
\definecolor{citecolor}{RGB}{0, 80, 200}
\definecolor{linkcolor}{RGB}{200, 80, 0}
\colorlet{colorours}{red!7}

\hypersetup{hidelinks}

\newtcolorbox{mycustombox}[1]{
  enhanced,
  colback=black!5!white,
  colframe=black!75!white,
  boxrule=0.4pt,
  coltitle=white,
  title=#1,
  titlerule=0.4pt,
  fontupper=\small,
  fonttitle=\small,
  before upper={\par\smallskipamount},
  breakable,
  left=3mm,
  right=3mm,
  top=2mm,
  bottom=2mm,
  boxsep=1mm,
}

\Crefformat{equation}{#2Eq.~(#1)#3}
\Crefformat{figure}{#2Fig.~#1#3}
\Crefformat{table}{#2Tab.~#1#3}
\Crefformat{assumption}{#2Assumption~#1#3}
\Crefformat{theorem}{#2Theorem~#1#3}
\Crefname{assumption}{Assumption}{Assumptions}
\Crefformat{section}{#2Sec.~#1#3}
\Crefformat{appendix}{#2Appendix~#1#3}
\Crefformat{algorithm}{#2Algorithm~#1#3}

\newcommand{\papertitle}{Roll Out and Roll Back: Diffusion LLMs are Their Own Efficiency Teachers}

\newcommand{\Method}{Wide-In, Narrow-Out}

\newcommand{\method}{WINO}

\theoremstyle{plain}

\theoremstyle{definition}

\theoremstyle{remark}

\makeatletter
\newcommand*\myfontsize{%
  \@setfontsize\myfontsize{7}{8}%
}
\makeatother
\newcommand{\mytextbox}[2]{\tikzmarknode[draw=#1,thick,inner sep=2pt]{test}{\myfontsize #2}}
\newcommand{\specialtoken}[2]{\mytextbox{#2}{\text{\textcolor{#2}{#1}}}}

\begin{document}
\title{\papertitle}

\author{
Fanqin Zeng$^*$,
Feng Hong$^*$,
Geng Yu,
Huangjie Zheng,
Xiaofeng Cao,
Ya Zhang,
Bo Han,
Yanfeng Wang,
Jiangchao Yao$^\dagger$
\thanks{Fanqin Zeng and Feng Hong contributed equally to this work (marked by $*$). The corresponding author is Jiangchao Yao (marked by $\dagger$), and the correspondence email is Sunarker@sjtu.edu.cn.}
\thanks{Fanqin Zeng, Feng Hong, Geng Yu, Ya Zhang, Yanfeng Wang, Jiangchao Yao are with Shanghai Jiao Tong University, Shanghai, China. Huangjie Zheng is with Apple MLR, Cupertino, CA, USA. Xiaofeng Cao is with Tongji University, Shanghai, China. Bo Han is with Hong Kong Baptist University, Hong Kong, China, and RIKEN, Japan. 
}}

\maketitle

\begin{abstract}
Diffusion Large Language Models (DLLMs) promise fast parallel generation, yet open-source DLLMs still face a severe quality--speed trade-off: accelerating decoding by revealing multiple tokens often causes substantial quality degradation. We attribute this dilemma to a train--inference mismatch amplified by irreversible decoding. While training reconstructs tokens from randomly corrupted states, efficient inference requires an adaptive denoising order, where easier tokens are revealed earlier and context-dependent ones are deferred. 
This view motivates two complementary methods: an inference-time method that makes parallel decoding revokable, and a training-time extension that distills the reliable order exposed by this revokable process.
Accordingly, we first propose {\Method} ({\method}), a training-free decoding algorithm that enables revokable parallel generation. {\method} aggressively drafts multiple tokens, verifies generated tokens with enriched global context, and re-masks unreliable ones for later refinement.
Building on this discovered order, we further introduce {\method}+, which injects the verified denoising trajectories produced by {\method} into model parameters, aligning training with efficient inference.
Experiments on LLaDA and MMaDA show that {\method} improves both quality and efficiency, while {\method}+ further strengthens this progression.
On GSM8K, {\method} improves accuracy from 73.24\% to 75.82\% with a 6.10$\times$ step reduction, and {\method}+ further achieves 76.58\% with a 6.83$\times$ reduction. On Flickr30K, {\method}+ reaches a 16.22$\times$ step reduction with improved CIDEr. These results demonstrate that DLLMs can serve as their own efficiency teachers by first discovering reliable denoising orders through revokable decoding and then learning to follow them for faster generation.
Code is available at \url{https://github.com/Feng-Hong/WINO-DLLM/tree/WINO-plus}.
\end{abstract}


\section{Introduction}
\IEEEPARstart{A}{utoregressive} (AR) large language models\cite{radford2018improving, radford2019language}, such as the GPT series\cite{openai2022chatgpt}, have shown impressive performance in a wide range of language tasks. However, their foundational token-by-token generation mechanism introduces inherent limitations, including severe inference latency, susceptibility to error propagation\cite{DBLP:journals/corr/abs-2310-12397,DBLP:journals/corr/abs-2310-08118}, and challenges in maintaining global coherence\cite{mei2025surveycontextengineeringlarge}. In response, Diffusion Large Language Models (DLLMs) have emerged as a compelling non-autoregressive alternative, architected to overcome these bottlenecks. By generating tokens simultaneously\cite{DBLP:conf/nips/LiTGLH22}, DLLMs theoretically enable massive inference acceleration, while their native bidirectional attention offers improved consistency. 
The immense potential of DLLMs has also been showcased by proprietary, closed-source systems (\emph{e.g.}, Mercury Coder\cite{inceptionlabs2025mercury} and Gemini Diffusion\cite{deepmind2025geminidiffusion}), which have demonstrated astonishing speeds exceeding 1,000 tokens per second, serving as a powerful proof-of-concept.

\begin{figure*}[t!]
\centering
\includegraphics[width=0.92\textwidth]{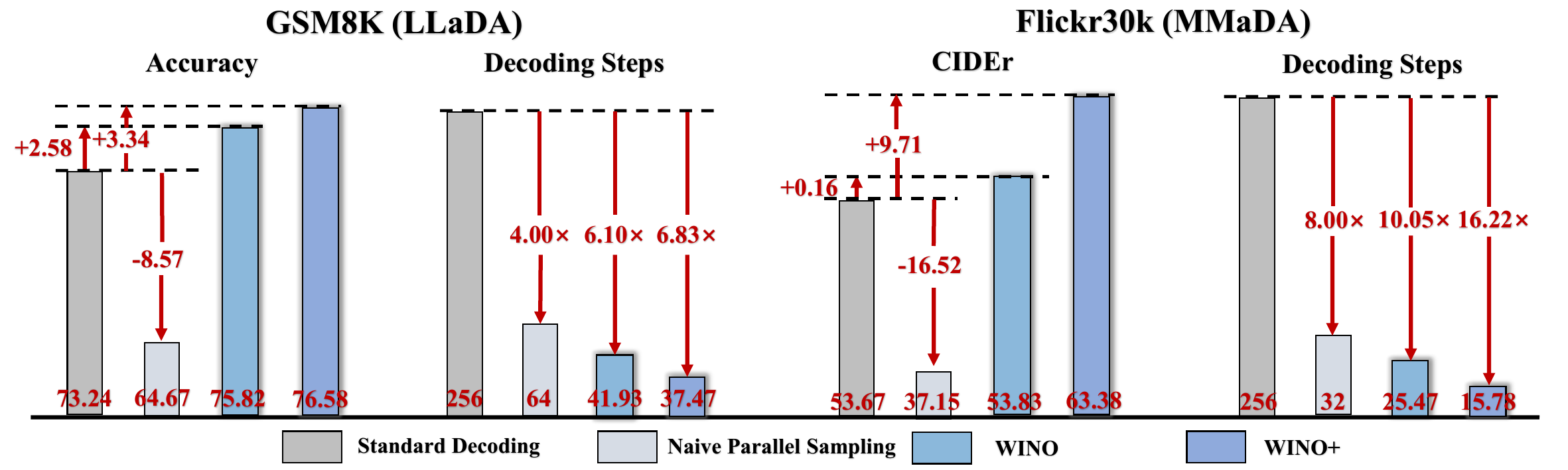}
\vspace{-10pt}
\caption{Demonstration of speedup and performance improvement of {\method} over standard decoding and naive parallel sampling evaluated on GSM8K with LLaDA and Flickr30K with MMaDA. The standard decoding unmasks 1 token per decoding step, while the naive parallel sampling unmasks $M(>1)$ tokens per decoding step. We set $M=4$ for GSM8K and $M=8$ for Flickr30K.}
\label{fig:speedup_demo}
\vspace{-10pt}
\end{figure*}

Despite this promise, the performance of open-source DLLMs has been still disappointing. One critical bottleneck is that they are caught in a severe quality-speed trade-off dilemma. Specifically, to achieve high-quality output, these models are often forced to decode slowly, generating just one token at a time, which negates their primary architectural advantage. As shown in Fig.~\ref{fig:speedup_demo}, attempting to accelerate inference by generating multiple tokens in parallel invariably leads to a significant degradation in output quality\cite{DBLP:journals/corr/abs-2502-09992,DBLP:journals/corr/abs-2506-00413}. 
This stark trade-off has largely prevented the open-source DLLMs from becoming a viable, high-performance alternative to their AR counterparts.

We attribute this trade-off to a train--inference mismatch in DLLMs, which is further amplified by the irreversibility of standard decoding~\cite{DBLP:conf/nips/SahooASGMCRK24, DBLP:conf/iclr/OuNXZSLL25}. During training, masked diffusion models reconstruct tokens from randomly corrupted states, where the recovery order is implicitly randomized. During inference, however, generation follows a progressive denoising trajectory, in which different tokens have different context requirements and thus a natural order preference. Easier tokens should be revealed earlier, while more context-dependent tokens should be delayed until sufficient context is available.
Standard decoding cannot correct violations of this order preference. 
Once a token is decoded, it is fixed and cannot be revised, even when richer context becomes available later. As a result, tokens revealed too early under insufficient context may introduce errors that are preserved and propagated throughout the remaining trajectory. Thus, the quality-speed trade-off stems from revealing tokens in a suboptimal order under an irreversible decoding process.

To address this problem from the inference side, we first propose 
{\Method} ({\method}), a novel decoding algorithm that enables revokable decoding for DLLMs. {\method} employs a draft-and-verify procedure that operates in parallel. At each step, a draft module aggressively proposes multiple new tokens based on a lenient threshold (the ``Wide-In''). Concurrently, a verify module leverages the newly enriched global context to re-evaluate all previously generated tokens. Any token that fails a stricter verification check is re-masked for refinement in a future step (the ``Narrow-Out''). 
This mechanism brings two merits: 
1) it breaks the irreversibility of conventional decoding, allowing tokens revealed too early to be revised when richer context becomes available; 
2) it permits more aggressive token generation in each diffusion step, yielding faster inference without sacrificing quality.
Importantly, WINO requires no additional training and can be directly applied to existing DLLMs.

While {\method} mitigates order-induced errors during inference, its draft--verify--fallback process also exposes a trajectory-level signal that can reduce the training-side mismatch. 
Each fallback indicates that a token has likely been revealed too early, while the finalization step of a token reflects a more suitable stage for revealing it.
 Based on this signal, we propose WINO+, a trajectory-injection framework that transfers the WINO-derived order into model parameters. WINO+ runs WINO offline, extracts token-level finalization steps, and constructs trajectory-guided training samples. Instead of reconstructing randomly selected masked tokens, WINO+ trains the model to reproduce the verified order: earlier-finalized tokens are supervised earlier, while later-finalized tokens remain masked until their corresponding trajectory step. This replaces random reconstruction with trajectory-ordered denoising, making training more aligned with efficient inference.

We conduct extensive experiments on both language and vision-language benchmarks using representative open-source DLLMs, including 
LLaDA\cite{DBLP:journals/corr/abs-2502-09992} and MMaDA\cite{DBLP:journals/corr/abs-2505-15809}. 
The results show a clear progression from WINO to WINO+. WINO accelerates decoding at inference time by making parallel generation revokable. WINO+ further improves efficiency from the training side by teaching the model to follow the verified generation order discovered by WINO, reducing the dependence on online fallback during inference.
For example, on GSM8K, WINO improves accuracy from 73.24\% to 75.82\% with a 6.10$\times$ reduction in decoding steps, while WINO+ further improves accuracy to 76.58\% with a 6.83$\times$ step reduction. On Flickr30K, WINO achieves a 10.05$\times$ step reduction, and WINO+ further improves CIDEr to 63.38 with a 16.22$\times$ step reduction. 
These results show that DLLMs can serve as their own efficiency teachers: their revokable decoding process first improves inference directly, and then produces verified trajectories from which the model can learn a better denoising order.
Our contributions are summarized as follows:
\begin{itemize}
    \item We identify a train--inference mismatch behind the quality-speed trade-off of DLLMs: standard training relies on randomly corrupted states, whereas efficient inference requires an adaptive denoising order. Under irreversible decoding, violations of this order preference lead to accumulated errors.
    
    \item We propose WINO, a training-free and plug-and-play decoding algorithm. Through a parallel draft--verify--fallback mechanism, WINO enables aggressive token generation while revising prematurely revealed tokens, thereby improving both decoding speed and generation quality.
    
    \item We further propose WINO+, a trajectory-injection framework that transfers the verified generation order discovered by WINO into model parameters. WINO+ extracts token-level finalization steps from WINO trajectories and trains the model with trajectory-ordered denoising instead of random reconstruction.

    \item Extensive experiments on language and vision-language benchmarks 
    show that WINO consistently accelerates inference while preserving or improving quality, and WINO+ further enhances both efficiency and performance by learning from WINO-derived trajectories.
\end{itemize}

\section{Related Work}

\subsection{Diffusion-based Language Models}
Diffusion models~\cite{DBLP:journals/corr/Sohl-DicksteinW15,DBLP:conf/nips/HoJA20,DBLP:conf/iclr/0011SKKEP21}, originally popularized in image generation~\cite{DBLP:conf/cvpr/RombachBLEO22,DBLP:conf/icml/NicholDRSMMSC22,DBLP:conf/nips/SahariaCSLWDGLA22}, have recently gained attention as an alternative to autoregressive language models (ARLMs) for text generation. Early diffusion work~\cite{DBLP:journals/corr/Sohl-DicksteinW15} first studied this expansion from continuous domain to discrete domain. Subsequently, D3PM~\cite{DBLP:conf/nips/AustinJHTB21} provides a general framework which models the diffusion forward process as a discrete state Markov chain defined by the multiplication of specific transition matrices over discrete time steps. A subsequent CTMC-based approach~\cite{DBLP:conf/nips/CampbellBBRDD22} later expands D3PM to a continuous time setting, utilizing the theory of continuous time Markov chain(CTMC).  
More recently, research on masked diffusion models(MDMs)
~\cite{DBLP:conf/nips/ShiHWDT24} derived from the absorbing state diffusion in D3PM has shown promising results both in small-scale models (\emph{e.g.}, MDLM~\cite{DBLP:conf/nips/SahooASGMCRK24} and RADD~\cite{DBLP:conf/iclr/OuNXZSLL25}) and large-scale implementations (\emph{e.g.}, LLaDA~\cite{DBLP:journals/corr/abs-2502-09992} and Dream~\cite{dream2025}). Extending this line of work, MMaDA~\cite{DBLP:journals/corr/abs-2505-15809} introduces a novel class of multimodal large diffusion models featuring a shared probabilistic formulation and a modality-agnostic architecture.

\subsection{DLLM Acceleration Techniques}
The existing acceleration study for DLLMs falls into two directions: KV cache and sampling compression. The former targets to build the KV cache for DLLMs due to its bidirectional full attention mechanism, unlike the causal attention of ARLMs. Typical works like Block Diffusion~\cite{DBLP:conf/iclr/ArriolaGCYQHSK25}, Fast-dLLM-cache~\cite{DBLP:journals/corr/abs-2505-22618} and dLLM-cache~\cite{DBLP:journals/corr/abs-2506-06295} respectively explore different caching mechanisms, which shows promising performance for speedup. Note that this direction is out of the scope of our work here. The latter direction focuses on optimizing the sampling process itself. For the classic low-confidence remasking strategy, several works have introduced novel sampling strategies to dynamically adjust the number of tokens predicted in parallel, thereby improving inference efficiency. Fast-dLLM-parallel~\cite{DBLP:journals/corr/abs-2505-22618} adopts a straightforward approach by selecting tokens with confidence scores exceeding a predefined threshold. Meanwhile, the entropy-bounded (EB) sampler~\cite{DBLP:journals/corr/abs-2505-24857}, as a drop-in replacement for conventional samplers, leverages an entropy-based unmasking procedure to dynamically decode multiple tokens per step while maintaining a predefined error tolerance. Although our {\method} brings the acceleration promise due to sampling compression, different from these works, we explore to address the inherent limitation of standard decoding in DLLMs.

\section{Preliminary: Decoding Process for DLLMs}

Given a prompt $X$, a DLLM generates a response $Y = [y_1, y_2, \ldots, y_L]$ with a pre-defined response length $L$. The response sequence is initialized as all special mask tokens, $Y^{(0)} = [\specialtoken{[MASK]}{gray},\specialtoken{[MASK]}{gray},\ldots,\specialtoken{[MASK]}{gray}]$. The decoding process iteratively refines the response sequence $Y^{(k)}$ over a total of $K$ denoising steps. In the following, we detail the case of $K = L$ (\emph{i.e.}, decoding one token per step), as existing models typically achieve optimal performance under this setting~\cite{DBLP:journals/corr/abs-2502-09992}. 

At step $k$, the goal of decoding is to refine the sequence $Y^{(k-1)}$ into $Y^{(k)}$. Given the token vocabulary $V$ and the model parameterized with $\theta$, the model estimates the probability distribution over the response sequence as $p_\theta(\hat{Y}|X, Y^{(k-1)})$. As a common example, in high-confidence greedy decoding, $Y^{(k)}$ is obtained by unmasking the most confident \specialtoken{[MASK]}{gray} token based on $Y^{(k-1)}$, \emph{i.e.},

\begin{equation}
\begin{array}{@{}l@{\;}c@{\;}l@{}}
l^{(k)}
&=&
\displaystyle
\mathop{\arg\max}_{l\in\{l|y^{(k-1)}_l=\specialtoken{[MASK]}{gray}\}}
\left(
\max_{v \in V} p_\theta(\hat{y}_l = v | X, Y^{(k-1)})
\right), \\[0.3em]
y^{(k)}_l
&=&
\begin{cases}
\argmax\limits_{v \in V} p_\theta(\hat{y}_l = v | X, Y^{(k-1)}),
& \text{if } l = l^{(k)}, \\
y^{(k-1)}_l,
& \text{otherwise},
\end{cases} \\[-0.1em]
&&
\hspace{0.25em}\forall l \in \{1,2,\dots,L\}.
\end{array}
\end{equation}

After completing all $K$ decoding steps, the final generated response is $Y = Y^{(K)}$. Existing DLLMs, such as LLaDA~\cite{DBLP:journals/corr/abs-2502-09992} and MMaDA~\cite{DBLP:journals/corr/abs-2505-15809}, can also accelerate the decoding process via naive parallel sampling by generating multiple tokens (\emph{e.g.}, 2 or 4) per step. However, empirical results reveal that such strategies often result in substantial performance degradation, limiting their practical effectiveness despite the computational speedup from using fewer decoding steps~\cite{DBLP:journals/corr/abs-2502-09992}.

\textbf{Semi-Autoregressive Diffusion Decoding.} This strategy is widely adopted by DLLMs like LLaDA~\cite{DBLP:journals/corr/abs-2502-09992} and MMaDA~\cite{DBLP:journals/corr/abs-2505-15809}, which involves splitting the response sequence into multiple blocks and decoding them sequentially from left to right. Within each block, the typical diffusion decoding strategy described above is applied.

We also provide the objective function for standard DLLM training in Appendix A as background reference.

\section{Roll Out: WINO Algorithm}
\label{sec:wino-algorithm}
\subsection{Key Limitation of Decoding Process for DLLMs}

While architecturally suited for parallelism, DLLMs still suffer from a severe quality-speed trade-off in effective multi-token decoding. As discussed in the Introduction, this trade-off stems from a train--inference mismatch: training reconstructs tokens from randomly corrupted states, whereas efficient inference requires an adaptive denoising order. During progressive denoising, different tokens have different context requirements. Aggressive parallel decoding may therefore reveal context-dependent tokens too early, before sufficient context is available to support reliable prediction.

This problem is further amplified by the irreversibility of standard decoding. Once a token is decoded, it is fixed and cannot be revised, even when richer context becomes available later. As a result, prematurely revealed tokens may be preserved and propagated through the remaining trajectory, degrading quality as decoding parallelism increases.

To address this problem from the inference side, we propose to break the irreversibility of standard decoding and introduce \emph{revokable decoding}. The key idea is to allow DLLMs to generate multiple tokens aggressively while retaining the ability to re-evaluate previously generated tokens under newly enriched global context. As more context emerges during generation, the model can correct its preliminary predictions. Such a mechanism effectively addresses the core conflict by marrying the efficiency of parallel generation with the accuracy of context-driven corrections.

\begin{figure*}[t]
    \centering
    \captionsetup{skip=2pt}
    \includegraphics[width=\textwidth, trim=0 10pt 0 0, clip]{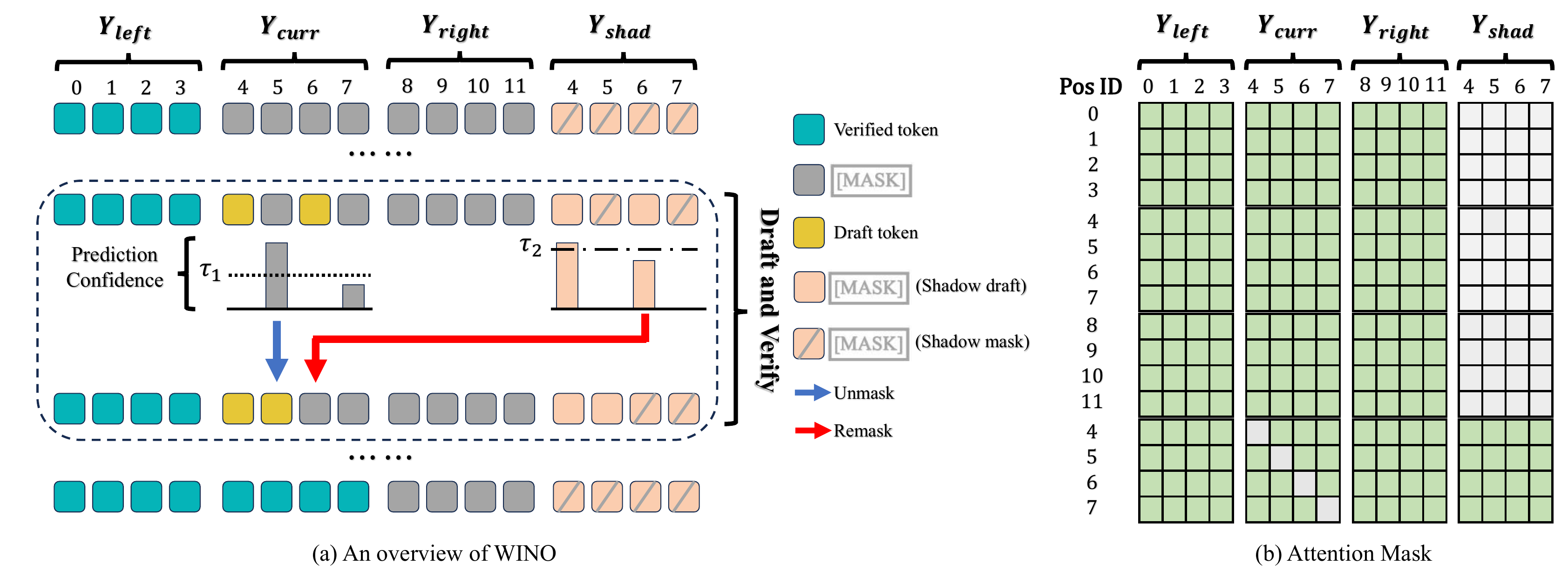}
    \caption{An overview of {\method} and illustration of our designed attention mask. The green squares denote 1, the grey squares denote 0, and ``Pos ID'' is short for position ID. Verified tokens refer to tokens in the prompt $X$ or previously decoded blocks. Draft tokens denote tokens in the current block that are unmasked up to the current decoding step. \specialtoken{[MASK]}{gray} (shadow draft) refer to tokens in the shadow block whose position IDs correspond to the draft tokens while \specialtoken{[MASK]}{gray} (shadow mask) refer to the remaining tokens in the shadow block.}
    \label{fig:method}
\end{figure*}

\subsection{Iterative Refinement via Parallel Draft-and-Verify}
Motivated by the above analysis and the design intuition, we propose a parallel Draft-and-Verify framework to enable revokable decoding for more efficient and higher-quality generation in DLLMs.

Specifically, our framework performs two modules in parallel at each decoding step: 1) Draft: aggressively unmasks multiple \specialtoken{[MASK]}{gray} tokens into candidate meaningful tokens; 2)	Verify: evaluates all currently unmasked tokens and re-masks those deemed low-quality for further refinement. We adopt the most common and general semi-autoregressive decoding paradigm to present our method. 
When the block length equals the generation length, this setting becomes equivalent to full diffusion decoding over the entire sequence.

\subsubsection{Drafting}
We denote the entire sequence as \( Y = [Y_{\text{left}}, {Y}_{\text{cur}}, Y_{\text{right}}] \), where \( Y_{\text{left}} \) contains the prompt \( X \) and the previously decoded blocks, \( {Y}_{\text{cur}} = [{y}_{\text{cur},1}, \ldots, {y}_{\text{cur},L_b}] \) represents the current block being decoded, and \( Y_\text{right} \) denotes the remaining blocks to be decoded. Here, \( L_b \) is the block length. At the $k$-th decoding step, instead of decoding a fixed number of tokens, we perform aggressive multi-token parallel decoding based on a confidence threshold $\tau_1$:
\begin{equation}
    {y}_{\text{cur},l}^{(k)} = \argmax\limits_{v \in V} p_\theta(\hat{{y}}_{\text{cur},l} = v\mid Y),
    \quad
    \text{if } \substack{\max\limits_{v \in V} p_\theta(\hat{{y}}_{\text{cur},l} = v\mid Y) > \tau_1 \\
    \text{and } {y}^{(k-1)}_{\text{cur},l}=\specialtoken{[MASK]}{gray}}.
\end{equation}

Here, a relatively low confidence threshold $\tau_1$ is adopted to allow more possible tokens to be decoded at each step, which will achieve the acceleration if only a few tokens among them are revoked during the verification module detailed in the next section. This will be demonstrated in the experiments.

\subsubsection{Verification}
The design principle of the verification module is to utilize the increasingly enriched semantic context at each decoding step—relative to earlier steps, to evaluate the quality of previously unmasked tokens. By re-masking low-quality tokens, the decoding process becomes revokable and amenable for the proper early error correction.

To realize effective quality verification about the decoded tokens, we design an auxiliary shadow block consisting entirely of \specialtoken{[MASK]}{gray}, $Y_{\text{shad}} = [\ \specialtoken{[MASK]}{gray}\ ]\times L_b$. This block is appended to the sequence $Y$, resulting in an extended sequence $\tilde{Y} = [Y_{\text{left}}, {Y}_{\text{cur}}, Y_{\text{right}},Y_{\text{shad}}]$. We carefully design the position IDs and attention mask associated with $Y_{\text{shad}}$ to ensure that its output can effectively verify the quality of the tokens decoded at the corresponding positions in $Y_{\text{cur}}$.

\textbf{Position IDs.} Although $Y_{\text{shad}}$ is appended to the right end of the sequence, we assign it the same position IDs as $Y_{\text{cur}}$. Thus, the output of $Y_{\text{shad}}$ corresponds to the same positions as $Y_{\text{cur}}$, enabling position-wise verification. 

\textbf{Attention Mask.} As illustrated in Fig.~\ref{fig:method}, we carefully design the attention mask after incorporating $Y_{\text{shad}}$ into the sequence $\tilde{Y}$. Specifically, tokens in $Y_{\text{left}}$, $Y_{\text{cur}}$, and $Y_{\text{right}}$ can freely attend to each other, but they are not allowed to attend to $Y_{\text{shad}}$. In contrast, each token in $Y_{\text{shad}}$ is allowed to attend to all tokens except its corresponding position in $Y_{\text{cur}}$. 

With the above design of position IDs and attention masks, we achieve the following properties:
\begin{itemize}
    \item For any token in the current block $Y_{\text{cur}}$, appending $Y_{\text{shad}}$ does not affect the model's output. Formally,
    $$p_\theta(\hat{{y}}_{\text{cur},l} |Y) = p_\theta(\hat{{y}}_{\text{cur},l} |\tilde{Y}).
    $$
    
    \item For any token in $Y_{\text{shad}}$, the following properties hold. For example, consider the token $y_{\text{shad},3}$ in Fig.~\ref{fig:method}, which is assigned position ID 6. 
    \begin{itemize}
	\item It shares the same position ID as $y_{\text{cur},3}$, and is allowed to attend to $Y_{\text{left}}$ and $Y_{\text{right}}$;
	\item It is explicitly prevented from attending to $y_{\text{cur},3}$, effectively avoiding information leakage during verification;
	\item For all other positions in $Y_{\text{cur}}$, each position is attended by exactly one decoded token (from $Y_{\text{cur}}$) and one \specialtoken{[MASK]}{gray} in $Y_{\text{shad}}$. The former provides progressively richer contextual semantics during decoding, while the latter serves to regularize the confidence of decoded tokens in $Y_{\text{cur}}$, reflecting the uncertainty and the need for potential refinement.
    \end{itemize}
\end{itemize}

With the specially designed position IDs and the attention mask described above, the verification module can be formally expressed as:
\begin{equation}
    {y}_{\text{cur},l}^{(k)} = \specialtoken{[MASK]}{gray},
    \quad
    \text{if } \substack{p_\theta(\hat{{y}}_{\text{shad},l} = {y}_{\text{cur},l}^{(k-1)}\mid \tilde{Y}) < \tau_2 \\
    \text{and } {y}_{\text{cur},l}^{(k-1)} \neq \specialtoken{[MASK]}{gray}}.
\end{equation}
where $\tau_2$ is the confidence threshold for verification.

\subsubsection{Overall Procedure}
In summary, at decoding step $k$, our framework enables both the drafting and verification processes to be completed in a single forward pass:
\begin{equation}
\small
\label{eq: overall}
    {y}_{\text{cur},l}^{(k)} =
\begin{cases}
\argmax\limits_{v \in V} p_\theta(\hat{{y}}_{\text{cur},l} = v\mid \tilde{Y}),
&\text{if } \substack{\max\limits_{v \in V} p_\theta(\hat{{y}}_{\text{cur},l} = v\mid \tilde{Y}) > \tau_1 \\
\text{and } {y}^{(k-1)}_{\text{cur},l}=\specialtoken{[MASK]}{gray}}, \\
\specialtoken{[MASK]}{gray},
&\text{if } \substack{p_\theta(\hat{{y}}_{\text{shad},l} = {y}_{\text{cur},l}^{(k-1)}\mid \tilde{Y}) < \tau_2 \\
\text{and } {y}_{\text{cur},l}^{(k-1)} \neq \specialtoken{[MASK]}{gray}}, \\
{y}_{\text{cur},l}^{(k-1)},
&\text{otherwise}.
\end{cases}
\end{equation}

 We iteratively refine the entire $Y_{\text{cur}}$ using the procedure in Eq.~(\ref{eq: overall}), until all tokens in $Y_{\text{cur}}$ are no longer \specialtoken{[MASK]}{gray}. We set the drafting threshold $\tau_1$ and the verification threshold $\tau_2$ such that $\tau_1 < \tau_2$. A lower $\tau_1$ accelerates the decoding process by allowing more tokens to be generated in parallel, while a higher $\tau_2$ ensures the quality of the final output by enforcing stricter acceptance criteria. We refer to this design philosophy as \emph{"Wide-In, Narrow-Out"}, abbreviated as {\method} in short.

Beyond online decoding, WINO also provides a reliable signal for training-time alignment. During draft and verification, a token may be revealed, revoked, and later revealed again under richer context. 

This behavior indicates that the first high-confidence prediction is not always the proper time to expose a token. 
Instead, the step from which a token stays stable reflects a more reliable generation order. 
In Section~\ref{sec:trajectory-injection}, we 
use this verified order to construct trajectory-guided training samples, so that the model can learn the denoising order discovered by WINO from its own.

\section{Roll Back: Learning from Verified WINO Trajectories}
\label{sec:trajectory-injection}

WINO improves DLLM inference by making parallel decoding revokable. 
However, its benefit is obtained online: the model still needs draft and verification to identify which tokens are safe to reveal. 
This suggests a natural training objective. 
If the model can learn the verified generation order discovered by WINO, it can reveal reliable tokens earlier and defer uncertain positions without relying heavily on online rollback.

We therefore propose WINO+, a trajectory guided training method that internalizes the verified denoising order produced by WINO. 
For each training instance, we first run WINO offline and record when each generated token becomes stable. 
We then construct training states that follow this stable order. 
The model is trained to predict the tokens scheduled for the current step, while keeping later tokens uncertain when they are not yet reliable. 
In this way, WINO+ aligns model training with the efficient inference process required by DLLMs.
The overall pipeline is illustrated in Fig.~\ref{fig:wino-plus-pipeline}.

\subsection{Verified Trajectories from WINO}
Given an input prompt $X$ and its ground truth $g$, WINO produces a denoising trajectory $\mathcal{T}^{W} = (Y^{(0)},Y^{(1)},\ldots,Y^{(K)})$, where $Y^{(0)}$ is fully masked on the response side and $Y^{(K)}$ is the final decoded response. We retain a trajectory only when its final response is judged correct by the task evaluator with respect to $g$. For example, in reasoning tasks, the extracted answer from $Y^{(K)}$ must match the ground truth answer; in code generation, the code extracted must pass the corresponding tests. For each retained trajectory, we use its final response as the training target and denote it by $Y^\dagger=[y^\dagger_1,\ldots,y^\dagger_L]$, namely $Y^\dagger=Y^{(K)}$.

For each response position $l$, we define its WINO finalization step as $t_l=\min \left\{t \mid \forall j\geq t,\; y_l^{(j)}=y_l^\dagger \right\}$. For prompt positions, we set $t_l=0$. Thus, $t_l$ records the earliest step after which position $l$ remains equal to its final generated token. A smaller $t_l$ means that the token can be reliably revealed earlier, while a larger $t_l$ means that the token requires more context before it becomes stable.
The offline trajectory-collection stage in \cref{fig:wino-plus-pipeline} illustrates how WINO rollouts are filtered and converted into token-level finalization steps.

The set of finalization steps gives a token wise generation order verified by WINO. 
Unlike random masking used in standard DLLM training, this order reflects the actual decoding process under aggressive parallel generation and verification. 
It therefore provides direct supervision for when each token should be revealed.

\subsection{Trajectory Guided Training States}

From each retained trajectory, we construct training states according to the extracted finalization steps. Let $\mathcal{S}=\{t_l \mid t_l>0\}$ be the set of response side finalization steps. For each $t\in\mathcal{S}$ we construct a partially revealed sequence $\widetilde{Y}^{(t-1)}$ by revealing only the tokens finalized before step $t$:
\begin{equation}
    \widetilde{y}^{(t-1)}_l=
\begin{cases}
y^\dagger_l, & t_l<t,\\
\texttt{[MASK]}, & t_l\geq t.
\end{cases}
\end{equation}
This state represents the context available immediately before the tokens finalized at step $t$ should be revealed.

\begin{figure*}[t]
    \centering
    \includegraphics[width=0.95\textwidth]{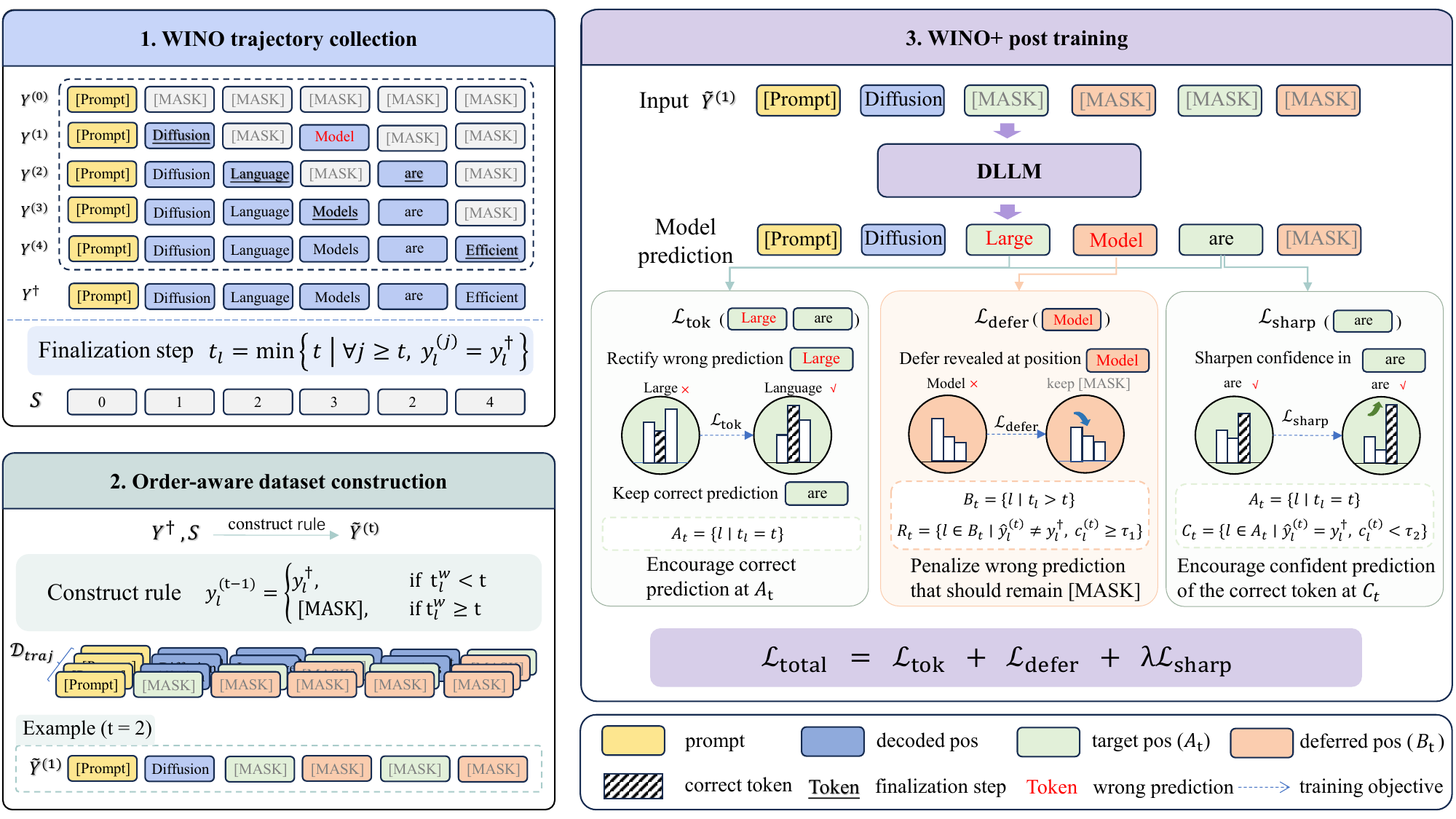}
    \caption{Overview of {\method}+. Stage 1 uses {\method} as an offline teacher to collect verification-guided trajectories and derive token finalization steps. Stage 2 converts the final response and its finalization-step vector into order-aware training states, where only tokens finalized before the current step are revealed. Stage 3 post-trains the base DLLM with trajectory-consistency losses: $\mathcal{L}_{\mathrm{tok}}$ supervises current-step targets, $\mathcal{L}_{\mathrm{defer}}$ suppresses premature high-confidence errors on deferred positions, and $\mathcal{L}_{\mathrm{sharp}}$ sharpens correct but low-confidence predictions.}
    \label{fig:wino-plus-pipeline}
\end{figure*}

We define $A_t=\{l\mid t_l=t\}$ as the positions to be revealed at the current step, and $B_t=\{l\mid t_l>t\}$ as the positions that should remain masked for later steps. Each tuple $(X,\widetilde{Y}^{(t-1)},A_t,B_t,Y^\dagger)$ is added to the trajectory guided dataset $\mathcal{D}_{\mathrm{traj}}$. This construction converts a WINO rollout into supervised training states that preserve the verified denoising order.
The order-aware dataset-construction stage in \cref{fig:wino-plus-pipeline} provides an example of how finalization steps are transformed into partially revealed training states.

\subsection{Trajectory Consistency Objective}

We train the base DLLM on $\mathcal{D}_{\mathrm{traj}}$. For a training tuple at step $t$, let $p_{\theta,l}^{(t)}(\cdot)=p_{\theta}(\cdot \mid X,\widetilde{Y}^{(t-1)})$ denote the prediction at position $l$. We use $\hat{y}^{(t)}_l$ and $c_l^{(t)}$ to denote its greedy prediction and confidence, respectively.

The primary goal is to make the model follow the verified order extracted from WINO. 
This gives a basic trajectory alignment objective with two complementary parts. First, positions in $A_t$ should be revealed at the current step, so we supervise them with cross entropy. Second, positions in $B_t$ should still be deferred. 
If the model produces a high confidence incorrect prediction on these deferred positions, the prediction may cause premature revealing during parallel decoding. 
We therefore suppress such predictions by maximizing their entropy.
Concretely, we define $R_t=\{l\in B_t \mid \hat{y}^{(t)}_l\neq y_l^\dagger,\; c_l^{(t)}\geq\tau_1\}$. The basic trajectory alignment loss is:
\begin{equation}
\small
    \mathcal{L}_{\mathrm{base}}^{(t)}
=
\underbrace{
-\frac{1}{|A_t|}
\sum_{l\in A_t}
\log p_{\theta,l}^{(t)}(y_l^\dagger)
}_{\mathcal{L}_{\mathrm{tok}}^{(t)}}
-
\underbrace{
\frac{1}{\max(1,|R_t|)}
\sum_{l\in R_t}
H\!\left(p_{\theta,l}^{(t)}\right)
}_{\mathcal{L}_{\mathrm{defer}}^{(t)}} .
\end{equation}
Here, $H(p)=-\sum_{v\in\mathcal{V}}p(v)\log p(v)$ denotes the Shannon entropy of the predictive distribution over the vocabulary $\mathcal{V}$. {$\mathcal{L}_{\mathrm{tok}}^{(t)}$ teaches what tokens should be revealed at the current verified step, while $\mathcal{L}_{\mathrm{defer}}^{(t)}$ discourages high confidence errors on tokens that should remain masked.}

Although this basic objective already transfers the WINO derived order, it does not explicitly improve the confidence of correct predictions that are close to being revealable. 
To further speed up inference, we add a confidence refinement term. 
Let $C_t=\{l\in A_t \mid \hat{y}^{(t)}_l=y_l^\dagger,\; c_l^{(t)}<\tau_2\}$ denote the positions whose predictions are correct but still below the verification threshold. 
We sharpen these predictions by minimizing their entropy, denoted as:
\begin{equation}
    \mathcal{L}_{\mathrm{sharp}}^{(t)}=\frac{1}{\max(1,|C_t|)}\sum_{l\in C_t}H(p_{\theta,l}^{(t)}).
\end{equation}

The final WINO+ objective is:
\begin{equation}
    \mathcal{L}^{(t)}
=
\mathcal{L}_{\mathrm{base}}^{(t)}
+
\lambda \mathcal{L}_{\mathrm{sharp}}^{(t)},
\end{equation}
where $\lambda$ is the sharp loss weight hyperparameter. The overall training objective is the expectation of $\mathcal{L}^{(t)}$ over all tuples in $\mathcal{D}_{\mathrm{traj}}$.

As shown in {\method}+ post training of Fig.~\ref{fig:wino-plus-pipeline},
this formulation separates the roles of the three losses. 
The token loss and the defer loss form the basic trajectory alignment objective with fixed unit weights. 
The sharp loss is used only as a confidence refinement term, controlled by \(\lambda\). 
Thus, WINO+ first learns the verified denoising order discovered by WINO, and then further sharpens correct current step predictions to enable faster token revealing during inference. 

At inference time, WINO+ uses the same semi-autoregressive decoding interface as the base DLLM. 
The auxiliary shadow block and online verification are not required. 
Because the model has been trained to follow the WINO derived order, more tokens can be safely revealed in each step, leading to faster decoding while preserving generation quality.

\begin{table*}[t]
\centering
\caption{Performance and inference speedup comparison on diverse language benchmarks.}
\label{tab:llada-bench}
\resizebox{0.98\textwidth}{!}{
\begin{tabular}{@{}!{\vrule width 0pt}p{2.4cm}<{\centering}p{1.2cm}<{\centering}p{2.5cm}<{\centering}p{2.5cm}<{\centering}p{1.8cm}<{\centering}p{2.5cm}<{\centering}p{1.8cm}<{\centering}}
\toprule
\multirow{2}{*}{\textbf{Benchmark}} & \multirow{2}{*}{\textbf{Method}} & \multirow{2}{*}{\textbf{Accuracy}} & \multirow{2}{*}{\textbf{Steps}}  & {\textbf{Step}} & \multirow{2}{*}{\textbf{TPS}} & {\textbf{TPS}} \\
&&&& \textbf{Reduction} && \textbf{Speedup}\\
\midrule

\multirow{3}{*}{\shortstack{GSM8K \\ \scriptsize \color{light-gray}{Math Reasoning}}}        & LLaDA & 73.24 & 256 & 1.00 $\times$ & 17.76 & 1.00 $\times$ \\
                              & \cellcolor{ourred} {\method}  & \cellcolor{ourred} 75.82 (\textcolor{ourblue}{+2.58}) & \cellcolor{ourred} 41.93 (\textcolor{ourblue}{-214.07}) & 
                                 \cellcolor{ourred} 6.10 $\times$ & 
                                 \cellcolor{ourred} 100.53 (\textcolor{ourblue}{+82.77}) & 
                                 \cellcolor{ourred} 5.66 $\times$ \\
                              & \cellcolor{colorours} {\method}+ & \cellcolor{colorours} 76.58 (\textcolor{ourblue}{+3.34}) & \cellcolor{colorours} 37.47 (\textcolor{ourblue}{-218.53}) & \cellcolor{colorours} 6.83 $\times$ & \cellcolor{colorours} 121.86 (\textcolor{ourblue}{+104.1}) & \cellcolor{colorours} 6.86 $\times$ \\
\midrule
\multirow{3}{*}{\shortstack{MATH-500 \\ \scriptsize \color{light-gray}{Math Reasoning}}}         & LLaDA & 32.00 & 256 & 1.00 $\times$ & 17.62 & 1.00 $\times$ \\
                              & \cellcolor{ourred} {\method}  & \cellcolor{ourred} 34.20 (\textcolor{ourblue}{+2.20}) & \cellcolor{ourred} 74.44 (\textcolor{ourblue}{-181.56}) & 
                                 \cellcolor{ourred} 3.44 $\times$ & 
                                 \cellcolor{ourred} 55.86 (\textcolor{ourblue}{+38.24}) & 
                                 \cellcolor{ourred} 3.17 $\times$ \\
                              & \cellcolor{colorours} {\method}+ & \cellcolor{colorours} 34.40 (\textcolor{ourblue}{+2.40}) & \cellcolor{colorours} 65.62 (\textcolor{ourblue}{-190.38}) & \cellcolor{colorours} 3.90 $\times$ & \cellcolor{colorours} 64.78 (\textcolor{ourblue}{+47.16}) & \cellcolor{colorours} 3.68 $\times$ \\
\midrule
\multirow{3}{*}{\shortstack{HumanEval \\ \scriptsize \color{light-gray}{Code Generation}}}   & LLaDA & 37.80 & 256 & 1.00 $\times$ & 14.52 & 1.00 $\times$ \\
                              & \cellcolor{ourred} {\method}  & \cellcolor{ourred} 42.07 (\textcolor{ourblue}{+4.27}) & \cellcolor{ourred} 93.32 (\textcolor{ourblue}{-162.68}) & 
                                 \cellcolor{ourred} 2.74 $\times$ & 
                                 \cellcolor{ourred} 37.19 (\textcolor{ourblue}{+22.67}) & 
                                 \cellcolor{ourred} 2.56 $\times$ \\
                              & \cellcolor{colorours} {\method}+ & \cellcolor{colorours} 42.68 (\textcolor{ourblue}{+4.88}) & \cellcolor{colorours} 82.71 (\textcolor{ourblue}{-173.29}) & \cellcolor{colorours} 3.10 $\times$ & \cellcolor{colorours} 37.80 (\textcolor{ourblue}{+23.28}) & \cellcolor{colorours} 2.60 $\times$ \\
\midrule
\multirow{3}{*}{\shortstack{MBPP \\ \scriptsize \color{light-gray}{Code Generation}}}        & LLaDA & 36.40 & 256 & 1.00 $\times$ & 18.52 & 1.00 $\times$ \\
                              & \cellcolor{ourred} {\method}  & \cellcolor{ourred} 36.40 (\textcolor{ourblue}{+0.00}) & \cellcolor{ourred} 96.57 (\textcolor{ourblue}{-159.43}) & 
                                 \cellcolor{ourred} 2.65 $\times$ & 
                                 \cellcolor{ourred} 45.39 (\textcolor{ourblue}{+26.87}) & 
                                 \cellcolor{ourred} 2.45 $\times$ \\
                              & \cellcolor{colorours} {\method}+ & \cellcolor{colorours} 37.20 (\textcolor{ourblue}{+0.80}) & \cellcolor{colorours} 67.53 (\textcolor{ourblue}{-188.47}) & \cellcolor{colorours} 3.79 $\times$ & \cellcolor{colorours} 64.75 (\textcolor{ourblue}{+46.23}) & \cellcolor{colorours} 3.50 $\times$ \\
\midrule
\multirow{3}{*}{\shortstack{Countdown \\ \scriptsize \color{light-gray}{Logical Reasoning}}} & LLaDA & 24.21 & 256 & 1.00 $\times$ & 17.22 & 1.00 $\times$ \\
                              & \cellcolor{ourred} {\method}  & \cellcolor{ourred} 33.20 (\textcolor{ourblue}{+8.99}) & \cellcolor{ourred} 105.88 (\textcolor{ourblue}{-150.12}) & 
                                 \cellcolor{ourred} 2.41 $\times$ & 
                                 \cellcolor{ourred} 38.97 (\textcolor{ourblue}{+21.75}) & 
                                 \cellcolor{ourred} 2.26 $\times$ \\
                              & \cellcolor{colorours} {\method}+ & \cellcolor{colorours} 48.05 (\textcolor{ourblue}{+23.84}) & \cellcolor{colorours} 63.47 (\textcolor{ourblue}{-192.53}) & \cellcolor{colorours} 4.03 $\times$ & \cellcolor{colorours} 71.48 (\textcolor{ourblue}{+54.26}) & \cellcolor{colorours} 4.15 $\times$ \\
\midrule
\multirow{3}{*}{\shortstack{Sudoku \\ \scriptsize \color{light-gray}{Logical Reasoning}}}    & LLaDA & 14.23 & 256 & 1.00 $\times$ & 11.61 & 1.00 $\times$ \\
                              & \cellcolor{ourred} {\method}  & \cellcolor{ourred} 15.20 (\textcolor{ourblue}{+0.97}) & \cellcolor{ourred} 131.96 (\textcolor{ourblue}{-124.04}) & 
                                 \cellcolor{ourred} 1.94 $\times$ & 
                                 \cellcolor{ourred} 21.11 (\textcolor{ourblue}{+9.50}) & 
                                 \cellcolor{ourred} 1.82 $\times$ \\
                              & \cellcolor{colorours} {\method}+ & \cellcolor{colorours} 18.37 (\textcolor{ourblue}{+4.14}) & \cellcolor{colorours} 64.77 (\textcolor{ourblue}{-191.23}) & \cellcolor{colorours} 3.95 $\times$ & \cellcolor{colorours} 48.85 (\textcolor{ourblue}{+37.24}) & \cellcolor{colorours} 4.21 $\times$ \\
\midrule
\multirow{3}{*}{\shortstack{ARC-E \\ \scriptsize \color{light-gray}{Commonsense Reasoning}}}            & LLaDA & 59.13 & 256 & 1.00 $\times$ & 17.26 & 1.00 $\times$ \\
                              & \cellcolor{ourred} {\method}  & \cellcolor{ourred} 81.19 (\textcolor{ourblue}{+22.06}) & \cellcolor{ourred} 40.19 (\textcolor{ourblue}{-215.81}) & 
                                 \cellcolor{ourred} 6.37 $\times$ & 
                                 \cellcolor{ourred} 101.61 (\textcolor{ourblue}{+84.35}) & 
                                 \cellcolor{ourred} 5.89 $\times$ \\
                              & \cellcolor{colorours} {\method}+ & \cellcolor{colorours} 84.97 (\textcolor{ourblue}{+25.84}) & \cellcolor{colorours} 24.86 (\textcolor{ourblue}{-231.14}) & \cellcolor{colorours} 10.30 $\times$ & \cellcolor{colorours} 178.75 (\textcolor{ourblue}{+161.49}) & \cellcolor{colorours} 10.36 $\times$ \\
\midrule
\multirow{3}{*}{\shortstack{ARC-C \\ \scriptsize \color{light-gray}{Commonsense Reasoning}}}            & LLaDA & 51.87 & 256 & 1.00 $\times$ & 17.10 & 1.00 $\times$ \\
                              & \cellcolor{ourred} {\method}  & \cellcolor{ourred} 73.89 (\textcolor{ourblue}{+22.02}) & \cellcolor{ourred} 47.41 (\textcolor{ourblue}{-208.59}) & 
                                 \cellcolor{ourred} 5.40 $\times$ & 
                                 \cellcolor{ourred} 85.42 (\textcolor{ourblue}{+68.32}) & 
                                 \cellcolor{ourred} 5.00 $\times$ \\
                              & \cellcolor{colorours} {\method}+ & \cellcolor{colorours} 80.60 (\textcolor{ourblue}{+28.73}) & \cellcolor{colorours} 28.78 (\textcolor{ourblue}{-227.22}) & \cellcolor{colorours} 8.90 $\times$ & \cellcolor{colorours} 150.61 (\textcolor{ourblue}{+133.51}) & \cellcolor{colorours} 8.81 $\times$ \\
\bottomrule
\end{tabular}}
\end{table*}

\section{Experiments}

\subsection{Experiment Setup}

\subsubsection{Datasets and Baselines} We conduct experiments to evaluate {\method} and {\method}+ across different types of tasks and domains. Specifically, for the language domain, we compare {\method} and {\method}+ with the standard decoding of LLaDA on eight tasks: GSM8K~\cite{DBLP:journals/corr/abs-2110-14168}, MATH-500~\cite{DBLP:conf/nips/HendrycksBKABTS21}, HumanEval~\cite{DBLP:journals/corr/abs-2107-03374}, MBPP~\cite{DBLP:journals/corr/abs-2108-07732}, Countdown~\cite{DBLP:journals/corr/abs-2504-12216}, Sudoku~\cite{DBLP:journals/corr/abs-2504-12216}, ARC-E~\cite{DBLP:journals/corr/abs-1803-05457}, and ARC-C~\cite{DBLP:journals/corr/abs-1803-05457}, covering four categories of textual generation tasks, including math reasoning, code generation, logical reasoning, and commonsense reasoning. For the vision-language domain, we evaluate {\method} and {\method}+ against the standard decoding of MMaDA~\cite{DBLP:journals/corr/abs-2505-15809} on six multimodal understanding tasks: Flickr30k~\cite{DBLP:journals/tacl/YoungLHH14}, AI2D~\cite{DBLP:journals/corr/KembhaviSKSHF16}, MATH-Vision~\cite{DBLP:conf/nips/WangPSLRZZL24}, MathVista~\cite{DBLP:conf/iclr/LuBX0LH0CG024}, MMMU~\cite{DBLP:conf/cvpr/YueNZ0LZSJRSWYY24}, and ScienceQA~\cite{DBLP:conf/nips/LuMX0CZTCK22}, spanning four types of multimodal tasks---captioning, chart understanding, math reasoning, and multi-discipline reasoning. For clarity, we test on the validation set of MMMU and the official testmini subset of MathVista.

\subsubsection{Evaluation Details}

All benchmarks are evaluated in a zero-shot manner, except Sudoku, which is evaluated in a 4-shot setting. 
We use CIDEr~\cite{DBLP:conf/cvpr/VedantamZP15} for Flickr30k and accuracy for all remaining benchmarks. 
To assess inference efficiency, we report the required decoding steps and Tokens Per Second (TPS) for the standard baselines, WINO, and WINO+, averaged over all samples in each benchmark. 
We adopt the open-sourced
\href{https://huggingface.co/GSAI-ML/LLaDA-8B-Instruct}{LLaDA-8B-Instruct}
for language benchmarks and
\href{https://huggingface.co/Gen-Verse/MMaDA-8B-MixCoT}{MMaDA-8B-MixCoT}
for vision-language tasks. 
For all evaluated decoding variants, including the standard baselines, WINO, and WINO+, we employ the semi-autoregressive sampling strategy introduced in LLaDA~\cite{DBLP:journals/corr/abs-2502-09992}, where the output sequence is partitioned into multiple blocks and generated from left to right. 
Unless specified otherwise, we set the generation length to 256 and the block length to 128 for all methods. 
For WINO, we set the verification threshold $\tau_{2}$ to 0.9 and tune the drafting threshold $\tau_{1}$ from $\{0.5, 0.6, 0.7\}$. 
For WINO+, we use the same generation and block-wise decoding configuration, but replace the draft-and-verify rule with a single-confidence-threshold parallel decoding strategy, where the confidence threshold is tuned from $[0.5, 0.9]$.

\subsubsection{WINO+ Training}

WINO+ is obtained by LoRA-based post-training~\cite{DBLP:conf/iclr/HuSWALWWC22} on WINO-guided trajectories. For LLaDA-8B-Instruct, we construct trajectories from GSM8K~\cite{DBLP:journals/corr/abs-2110-14168} and Countdown-Tasks-3to4~\cite{DBLP:journals/corr/abs-2504-12216}; for MMaDA-8B-MixCoT, we use IconQA~\cite{DBLP:conf/nips/LuQCXZZYLZ21}. LoRA adapters are inserted into the attention and MLP projection layers, with rank and scaling factor set to 128. Training is conducted with bf16 mixed precision and AdamW optimization~\cite{loshchilov2019decoupled}. Full details on data construction, trajectory filtering, loss weights, and model-specific optimization settings are provided in Appendix B.

\subsection{Main Results}

\begin{table*}[t]
\centering
\caption{Performance and inference speedup comparison across diverse multi-modal understanding and reasoning benchmarks. We use CIDEr for Flickr30k and accuracy for other benchmarks.}
\label{tab:mmada-bench}
\resizebox{0.98\textwidth}{!}{
\begin{tabular}{@{}!{\vrule width 0pt}p{2.5cm}<{\centering}p{1.2cm}<{\centering}p{2.5cm}<{\centering}p{2.5cm}<{\centering}p{1.8cm}<{\centering}p{2.5cm}<{\centering}p{1.8cm}<{\centering}}
\toprule
\multirow{2}{*}{\textbf{Benchmark}} & \multirow{2}{*}{\textbf{Method}} & \multirow{2}{*}{\textbf{Performance}} & \multirow{2}{*}{\textbf{Steps}}  & {\textbf{Step}} & \multirow{2}{*}{\textbf{TPS}} & {\textbf{TPS}} \\
&&&& \textbf{Reduction} && \textbf{Speedup}\\
\midrule 
\multirow{3}{*}{\shortstack{Flickr30k \\ \scriptsize \color{light-gray}{Captioning}}}                  & MMaDA & 53.67 & 256 & 1.00 $\times$ & 6.41 & 1.00 $\times$ \\
       & \cellcolor{ourred} {\method}  & \cellcolor{ourred} 53.83 (\textcolor{ourblue}{+0.16}) & \cellcolor{ourred} 25.47 (\textcolor{ourblue}{-230.53}) & 
                                 \cellcolor{ourred} 10.05 $\times$ & 
                                 \cellcolor{ourred} 55.11 (\textcolor{ourblue}{+48.70}) & 
                                 \cellcolor{ourred} 8.60 $\times$ \\
                              & \cellcolor{colorours} {\method}+ & \cellcolor{colorours} 63.38 (\textcolor{ourblue}{+9.71}) & \cellcolor{colorours} 15.78 (\textcolor{ourblue}{-240.22}) & \cellcolor{colorours} 16.22 $\times$ & \cellcolor{colorours} 106.07 (\textcolor{ourblue}{+99.66}) & \cellcolor{colorours} 16.55 $\times$ \\
\midrule
\multirow{3}{*}{\shortstack{AI2D \\ \scriptsize \color{light-gray}{Chart Understanding}}}              & MMaDA & 54.86 & 256 & 1.00 $\times$ & 6.31 & 1.00 $\times$  \\
        & \cellcolor{ourred} {\method}  & \cellcolor{ourred} 57.19 (\textcolor{ourblue}{+2.33}) & \cellcolor{ourred} 30.90 (\textcolor{ourblue}{-225.10}) & 
                                 \cellcolor{ourred} 8.30 $\times$ & 
                                 \cellcolor{ourred} 46.04 (\textcolor{ourblue}{+39.73}) & 
                                 \cellcolor{ourred} 7.30 $\times$ \\
                              & \cellcolor{colorours} {\method}+ & \cellcolor{colorours} 66.61 (\textcolor{ourblue}{+11.75}) & \cellcolor{colorours} 26.15 (\textcolor{ourblue}{-229.85}) & \cellcolor{colorours} 9.79 $\times$ & \cellcolor{colorours} 62.00 (\textcolor{ourblue}{+55.69}) & \cellcolor{colorours} 9.83 $\times$ \\
\midrule
\multirow{3}{*}{\shortstack{MATH-Vision \\ \scriptsize \color{light-gray}{Math Reasoning}}}            & MMaDA & 8.55  & 256 & 1.00 $\times$ & 6.22 & 1.00 $\times$ \\
        & \cellcolor{ourred} {\method}  & \cellcolor{ourred} 9.57 (\textcolor{ourblue}{+1.02}) & \cellcolor{ourred} 44.69 (\textcolor{ourblue}{-211.31}) & 
                                 \cellcolor{ourred} 5.73 $\times$ & 
                                 \cellcolor{ourred} 31.17 (\textcolor{ourblue}{+24.95}) & 
                                 \cellcolor{ourred} 5.01 $\times$ \\
                              & \cellcolor{colorours} {\method}+ & \cellcolor{colorours} 13.42 (\textcolor{ourblue}{+4.87}) & \cellcolor{colorours} 41.06 (\textcolor{ourblue}{-214.94}) & \cellcolor{colorours} 6.23 $\times$ & \cellcolor{colorours} 37.46 (\textcolor{ourblue}{+31.24}) & \cellcolor{colorours} 6.02 $\times$ \\
\midrule
\multirow{3}{*}{\shortstack{MathVista-mini \\ \scriptsize \color{light-gray}{Math Reasoning}}}         & MMaDA & 31.10 & 256 & 1.00 $\times$ & 6.21 & 1.00 $\times$ \\
        & \cellcolor{ourred} {\method}  & \cellcolor{ourred} 31.40 (\textcolor{ourblue}{+0.30}) & \cellcolor{ourred} 33.45 (\textcolor{ourblue}{-222.55}) & 
                                 \cellcolor{ourred} 7.65 $\times$ & 
                                 \cellcolor{ourred} 41.96 (\textcolor{ourblue}{+35.75}) & 
                                 \cellcolor{ourred} 6.76 $\times$ \\
                              & \cellcolor{colorours} {\method}+ & \cellcolor{colorours} 31.40 (\textcolor{ourblue}{+0.30}) & \cellcolor{colorours} 19.41 (\textcolor{ourblue}{-236.59}) & \cellcolor{colorours} 13.19 $\times$ & \cellcolor{colorours} 82.16 (\textcolor{ourblue}{+75.95}) & \cellcolor{colorours} 13.23 $\times$ \\
\midrule
\multirow{3}{*}{\shortstack{MMMU-val \\ \scriptsize \color{light-gray}{Multi-discipline Reasoning}}}   & MMaDA & 18.56 & 256 & 1.00 $\times$ & 6.02 & 1.00 $\times$ \\
        & \cellcolor{ourred} {\method}  & \cellcolor{ourred} 24.00 (\textcolor{ourblue}{+5.44}) & \cellcolor{ourred} 38.47 (\textcolor{ourblue}{-217.53}) & 
                                 \cellcolor{ourred} 6.65 $\times$ & 
                                 \cellcolor{ourred} 36.13 (\textcolor{ourblue}{+30.11}) & 
                                 \cellcolor{ourred} 6.00 $\times$ \\
                              & \cellcolor{colorours} {\method}+ & \cellcolor{colorours} 28.11 (\textcolor{ourblue}{+9.55}) & \cellcolor{colorours} 26.25 (\textcolor{ourblue}{-229.75}) & \cellcolor{colorours} 9.75 $\times$ & \cellcolor{colorours} 54.18 (\textcolor{ourblue}{+48.16}) & \cellcolor{colorours} 9.00 $\times$ \\
\midrule
\multirow{3}{*}{\shortstack{ScienceQA \\ \scriptsize \color{light-gray}{Multi-discipline Reasoning}}}  & MMaDA & 30.89 & 256 & 1.00 $\times$ & 6.07 & 1.00 $\times$ \\
        & \cellcolor{ourred} {\method}  & \cellcolor{ourred} 42.24 (\textcolor{ourblue}{+11.35}) & \cellcolor{ourred} 28.12 (\textcolor{ourblue}{-227.88}) & 
                                 \cellcolor{ourred} 9.10 $\times$ & 
                                 \cellcolor{ourred} 49.45 (\textcolor{ourblue}{+43.38}) & 
                                 \cellcolor{ourred} 8.15 $\times$ \\
                              & \cellcolor{colorours} {\method}+ & \cellcolor{colorours} 53.84 (\textcolor{ourblue}{+22.95}) & \cellcolor{colorours} 23.26 (\textcolor{ourblue}{-232.74}) & \cellcolor{colorours} 11.01 $\times$ & \cellcolor{colorours} 66.89 (\textcolor{ourblue}{+60.82}) & \cellcolor{colorours} 11.02 $\times$ \\
\bottomrule
\end{tabular}}
\end{table*}

\subsubsection{Performance and speedup on text generation}

We report the performance, decoding steps and TPS of LLaDA, {\method}, and {\method}+ on language benchmarks in Table~\ref{tab:llada-bench}. Compared with standard LLaDA decoding, {\method} achieves better accuracy with far fewer decoding steps on most benchmarks, except for MBPP where it matches the baseline performance. For instance, {\method} improves GSM8K accuracy by 2.58\%, while reducing the average decoding steps from 256 to 41.93, corresponding to a 6.10$\times$ step reduction and a 5.66$\times$ TPS speedup. Similar trends can also be observed on MATH-500, HumanEval, Countdown, Sudoku, ARC-E, and ARC-C, showing that the draft-and-verify mechanism can improve generation quality while preserving the parallel decoding advantage of DLLMs. {\method}+ further improves the trade-off by injecting WINO-guided trajectories into the model. It achieves higher accuracy than both LLaDA and {\method} on all eight textual tasks, while requiring fewer decoding steps than {\method} on every benchmark. Notably, {\method}+ improves Countdown from 24.21\% to 48.05\%, ARC-E from 59.13\% to 84.97\%, and ARC-C from 51.87\% to 80.60\%, with 4.15$\times$, 10.36$\times$, and 8.81$\times$ TPS speedups, respectively. These results demonstrate that the draft-and-verify mechanism improves generation quality and inference efficiency, and that its induced trajectories can be further internalized by post-training.

\subsubsection{Performance and speedup on multimodal understanding and reasoning}

We further assess the performance and efficiency gains of {\method} and {\method}+ on multimodal benchmarks in Table~\ref{tab:mmada-bench}. Compared with vanilla MMaDA, {\method} achieves consistent and substantial speedup across all benchmarks. For example, it obtains a 10.05$\times$ step reduction and an 8.60$\times$ TPS speedup on Flickr30k while maintaining comparable captioning quality. On AI2D, MMMU-val, and ScienceQA, {\method} improves performance by 2.33, 5.44, and 11.35 points, respectively, with large efficiency gains. {\method}+ further strengthens this trend. It improves Flickr30k CIDEr from 53.67 to 63.38 with a 16.55$\times$ TPS speedup, and improves ScienceQA accuracy by 22.95 points with an 11.02$\times$ TPS speedup. On MMMU-val, it improves accuracy from 18.56\% to 28.11\% while reducing the average decoding steps to 26.25. These results show that revocable parallel decoding also benefits multimodal DLLMs, where visual and textual contexts jointly affect token reliability. The further gains of {\method}+ indicate that the verified trajectories discovered by {\method} can serve as effective supervision, helping the model internalize token-level generation order for a better quality--efficiency trade-off.

\begin{table*}[t]
\centering
\caption{Experiment results on different generation lengths and full diffusion setting, respectively. }
\label{tab:gen_length&full_diff}
\resizebox{0.98\textwidth}{!}{
\begin{tabular}{@{}!{\vrule width 0pt}p{1.5cm}<{\centering}p{1.5cm}<{\centering}p{1.5cm}<{\centering}p{1.2cm}<{\centering}p{2.5cm}<{\centering}p{1.5cm}<{\centering}p{1.5cm}<{\centering}p{1.5cm}<{\centering}p{1.5cm}<{\centering}}
\toprule
\multirow{2}{*}{\textbf{Benchmark}} & \textbf{Generation} & \textbf{Block} & \multirow{2}{*}{\textbf{Method}} & \multirow{2}{*}{\textbf{Accuracy}} & \multirow{2}{*}{\textbf{Steps}}  & {\textbf{Step}} & \multirow{2}{*}{\textbf{TPS}} & {\textbf{TPS}} \\
& \textbf{Length} & \textbf{Length} & & & & \textbf{Reduction} & & \textbf{Speedup}\\
\midrule
\multicolumn{9}{c}{\emph{Different Generation Lengths}} \\
\multirow{4}{*}{GSM8K}               & \multirow{2}{*}{256} & \multirow{2}{*}{128} & LLaDA     & 73.24 & 256   & 1.00 $\times$ & 17.76  & 1.00 $\times$        \\
                                     &                      & & \cellcolor{ourred} {\method} & \cellcolor{ourred} 75.82 (\textcolor{ourblue}{+2.58}) & \cellcolor{ourred} 41.93 &\cellcolor{ourred} 6.10 $\times$ &\cellcolor{ourred} 100.53 &\cellcolor{ourred} 5.66 $\times$      \\
                                     & \multirow{2}{*}{512} & \multirow{2}{*}{128} & LLaDA     & 74.60 & 512   & 1.00 $\times$ & 11.84 & 1.00 $\times$     \\
                                     &                      & & \cellcolor{ourred} {\method} &\cellcolor{ourred} 79.91 (\textcolor{ourblue}{+5.31}) &\cellcolor{ourred} 68.53 &\cellcolor{ourred} 7.47 $\times$ &\cellcolor{ourred} 82.64 &\cellcolor{ourred} 6.98 $\times$     \\
\midrule
\multirow{4}{*}{MMMU-val}            & \multirow{2}{*}{256} & \multirow{2}{*}{128} & MMaDA     & 18.56 & 256   & 1.00 $\times$ & 6.02  & 1.00 $\times$ \\
                                     &                      & &\cellcolor{ourred} {\method} &\cellcolor{ourred} 24.00 (\textcolor{ourblue}{+5.44}) &\cellcolor{ourred} 38.47 &\cellcolor{ourred} 6.65 $\times$ &\cellcolor{ourred} 36.13 &\cellcolor{ourred} 6.00 $\times$ \\
                                     & \multirow{2}{*}{512} & \multirow{2}{*}{128} & MMaDA     & 18.44 & 512 & 1.00 $\times$ & 5.01 & 1.00 $\times$     \\
                                     &                      & &\cellcolor{ourred} {\method} &\cellcolor{ourred} 23.44 
                                     (\textcolor{ourblue}{+5.00}) &\cellcolor{ourred} 64.82 &\cellcolor{ourred} 7.90 $\times$ &\cellcolor{ourred} 35.01 &\cellcolor{ourred} 6.99 $\times$  \\
\midrule\midrule
\multicolumn{9}{c}{\emph{Full Diffusion}} \\
\multirow{4}{*}{GSM8K}               & \multirow{2}{*}{256}  & \multirow{2}{*}{256} & LLaDA     & 34.34 & 256   & 1.00 $\times$ & 17.73 & 1.00 $\times$      \\
                                     &                       & &\cellcolor{ourred} {\method}   &\cellcolor{ourred} 58.22 (\textcolor{ourblue}{+23.88})  &\cellcolor{ourred} 38.77 &\cellcolor{ourred} 6.60 $\times$ &\cellcolor{ourred} 93.61 &\cellcolor{ourred} 5.28 $\times$  \\
                                     & \multirow{2}{*}{128}  & \multirow{2}{*}{128} & LLaDA     & 58.60 & 128   & 1.00 $\times$ & 23.23 & 1.00 $\times$  \\
                                     &                       & &\cellcolor{ourred} {\method}   &\cellcolor{ourred} 62.32 (\textcolor{ourblue}{+3.72}) &\cellcolor{ourred} 23.95 &\cellcolor{ourred} 5.34 $\times$ &\cellcolor{ourred} 114.29 &\cellcolor{ourred} 4.92 $\times$  \\
\midrule
\multirow{4}{*}{MMMU-val}            & \multirow{2}{*}{256}  & \multirow{2}{*}{256} & MMaDA     & 17.22 & 256   & 1.00$\times$ & 6.11 & 1.00$\times$   \\
                                     &                       & &\cellcolor{ourred} {\method} &\cellcolor{ourred} 22.44 (\textcolor{ourblue}{+5.22}) &\cellcolor{ourred} 24.94 &\cellcolor{ourred} 10.26$\times$ &\cellcolor{ourred} 50.03 &\cellcolor{ourred} 8.19$\times$    \\
                                     & \multirow{2}{*}{128}  & \multirow{2}{*}{128} & MMaDA     & 15.33 & 128   & 1.00$\times$  & 6.70 & 1.00 $\times$    \\
                                     &                       & & \cellcolor{ourred} \method &\cellcolor{ourred} 23.11 (\textcolor{ourblue}{+7.78}) &\cellcolor{ourred} 19.14 &\cellcolor{ourred} 6.69$\times$  &\cellcolor{ourred} 39.94 &\cellcolor{ourred} 5.96 $\times$    \\
\bottomrule
\end{tabular}
}
\end{table*}

\begin{table*}[t]
\centering
\caption{Experiment results on the variant of {\method} without the verification module.}
\label{tab:verification_module}
\resizebox{0.98\textwidth}{!}{
\begin{tabular}{@{}!{\vrule width 0pt}p{2.5cm}<{\centering}p{3.5cm}<{\centering}p{1.5cm}<{\centering}p{1.5cm}<{\centering}p{1.5cm}<{\centering}p{1.5cm}<{\centering}p{1.5cm}<{\centering}}
\toprule
\multirow{2}{*}{\textbf{Benchmark}} & \multirow{2}{*}{\textbf{Method}} & \multirow{2}{*}{\textbf{Accuracy}} & \multirow{2}{*}{\textbf{Steps}}  & {\textbf{Step}} & \multirow{2}{*}{\textbf{TPS}} & {\textbf{TPS}} \\
& & & & \textbf{Reduction} & & \textbf{Speedup}\\
\midrule
\multirow{4}{*}{GSM8K}      & LLaDA   & 73.24 & 256 & 1.00 $\times$ & 17.76 & 1.00 $\times$ \\     
                            & Only Draft ($\tau_1=0.6$) & 70.28 & 34.79 & 7.36 $\times$ & 130.89 & 7.37 $\times$       \\
                            & Only Draft ($\tau_1=0.9$) & 72.33 & 81.39 & 3.15 $\times$ & 56.12 & 3.16 $\times$       \\
                            & \method                   & 75.82 & 41.93 & 6.10 $\times$ & 100.53 & 5.66 $\times$      \\
\midrule
\multirow{4}{*}{MMMU-val}   & MMaDA & 18.56 & 256 & 1.00 $\times$ & 6.02 & 1.00 $\times$ \\
                            & Only Draft ($\tau_1=0.6$) & 19.89 & 35.63 & 7.18 $\times$ & 43.22 & 7.18 $\times$      \\
                            & Only Draft ($\tau_1=0.9$) & 18.56 & 79.74 & 3.21 $\times$ & 19.38 & 3.22 $\times$       \\
                            & \method                   & 24.00 & 38.47 & 6.65 $\times$ & 36.13 & 6.00 $\times$ \\
\bottomrule
\end{tabular}
}
\end{table*}

\subsection{Analysis and Ablation Studies of WINO}

\subsubsection{On different generation length}
In Table~\ref{tab:gen_length&full_diff}, we evaluate the performance of {\method} with experiments on different generation lengths, where the block length $L_b$ is fixed to 128 and the baselines unmask 1 token every decoding step (to achieve their best generation performance). When the generation length is set to 512, {\method} still achieves comparable or better task performance with significantly fewer decoding steps, demonstrating the effectiveness of {\method} across different generation lengths.

\subsubsection{On full diffusion decoding (instead of semi-autoregressive decoding)}
In Table~\ref{tab:gen_length&full_diff}, we compare the baselines and {\method} applying full diffusion decoding, which means the block length $L_b$ is set equal to the generation length. Compared to results on the semi-autoregressive decoding in Table~\ref{tab:llada-bench} and Table~\ref{tab:mmada-bench}, {\method} demonstrates substantially strong accuracy gains under the full diffusion setting. Notably, while LLaDA suffers a substantial accuracy drop on GSM8K with full diffusion decoding, {\method} maintains reasonable performance with far fewer decoding steps. These results indicate that {\method} unlocks significantly greater potential for boosting model performance and computational efficiency when applied in full diffusion decoding scenarios.

\subsubsection{Comparison with naive parallel sampling}
The decoding process of existing DLLMs can be sped up by generating multiple tokens per step, \emph{i.e.}, naive parallel sampling. However, directly increasing the fixed number of generated tokens per step for DLLMs leads to significant performance degradation. For instance, on GSM8K, accuracy drops from 73.24\% with 256 steps (1 token/step) to 71.11\% with 128 steps (2 tokens/step), and further down to 64.67\% with 64 steps (4 tokens/step). In contrast, the draft-and-verify procedure of {\method} enables flexible decoding during the generation process, achieving 75.82\% accuracy with only 41.93 steps on average, corresponding to a 6.10$\times$ speedup, thereby substantially improving task performance while accelerating inference.

\subsubsection{Ablation on verification module}
We conduct an ablation study on a variant of {\method} that excludes the verification module, implemented by setting the verification threshold $\tau_2$ to zero. As presented in Table~\ref{tab:verification_module}, this variant exhibits significant performance degradation across both benchmarks compared to {\method}. Specifically, when the drafting threshold $\tau_1$ is small (corresponding to 0.6 in the table), more candidate tokens are unmasked per decoding step, which naturally introduces a higher proportion of unreliable tokens and ultimately compromises output quality. Conversely, when $\tau_1$ is large (corresponding to 0.9 in the table), fewer candidate tokens are unmasked per decoding step, which can mitigate error propagation but at the expense of computational efficiency. Crucially, without the verification module, the generation process lacks a mechanism to correct erroneous predictions. As a result, even with a large $\tau_1$, the model may fail to achieve comparable performance, underscoring the necessity of the verification module in maintaining generation quality.

\begin{figure*}[!t]
\centering
\includegraphics[width=0.95\textwidth]{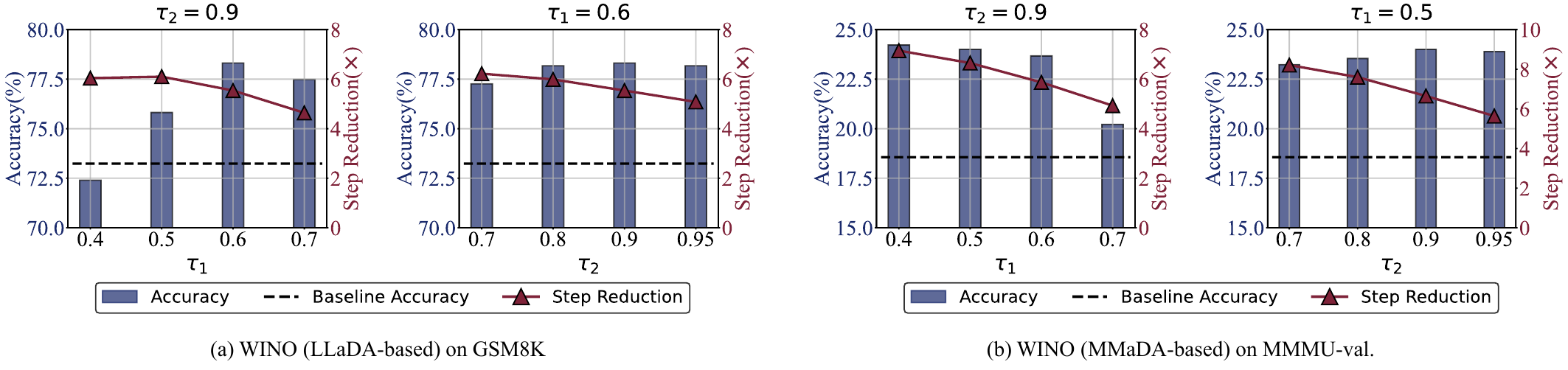}
\caption{Ablation of the drafting threshold $\tau_1$ and verification threshold $\tau_2$ in {\method}. Results are shown for LLaDA-based {\method} on GSM8K and MMaDA-based {\method} on MMMU-val; each plot varies one threshold while fixing the other, and reports accuracy together with step-reduction ratio.}
\label{fig:ab_tuning_tau}
\end{figure*}

\subsubsection{Effect of threshold tuning}
In Fig.~\ref{fig:ab_tuning_tau}, we present the evaluation results of {\method} with varying drafting threshold $\tau_1$ and verification threshold $\tau_2$. Our experiments suggest that {\method} consistently outperforms baselines across different benchmarks and the $\tau_1$ and $\tau_2$ values in terms of both task performance and inference efficiency. As the $\tau_1$ value decreases, more candidate tokens are unmasked at each decoding step, thereby accelerating inference by reducing the required decoding steps. However, this comes at the cost of introducing more unreliable predictions, which may place a greater burden on the verification module to correct errors. Empirically, we find that setting the value of drafting threshold $\tau_1$ within the range of 0.5 to 0.7 achieves an optimal balance, maintaining competitive task performance while preserving efficient generation. The verification threshold $\tau_2$ controls the strictness of the verification process and thus influences decoding speed. Since the performance is relatively robust to $\tau_2$, we fix $\tau_2 = 0.9$ in all experiments, while leaving open the possibility of further tuning this parameter for even better performance and speedup.

\subsubsection{Relation between speedup and task complexity}
\begin{figure}[t]
    \centering
    \includegraphics[width=0.8\columnwidth]{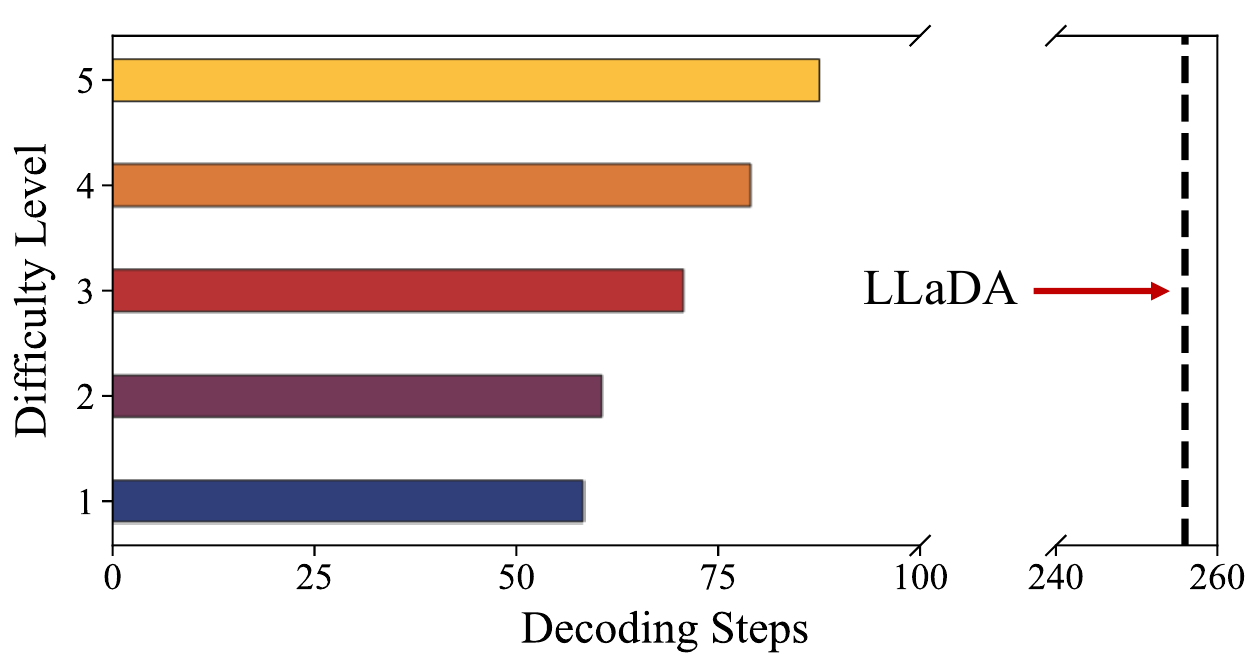}
    \vspace{-10pt}
    \caption{Decoding steps of {\method} on subsets of the MATH benchmark with varied difficulty levels.}
    \vspace{-5pt}
    \label{fig:difficulty_level}
\end{figure}

As shown in Table~\ref{tab:llada-bench} and Table~\ref{tab:mmada-bench}, we observe a consistent positive correlation between the degree of speedup and task performance across all benchmarks. For instance, {\method} achieves a 10.05$\times$ step reduction on the relatively simple captioning task Flickr30k, compared to only 5.73$\times$ step reduction on the more challenging math reasoning benchmark MATH-Vision. This is because models can, in principle, solve tasks they are more proficient at with lower computational cost, leaving greater room for acceleration under our decoding method. And since models are typically more confident when handling easier tasks, each decoding step in {\method} tends to yield a larger number of effective tokens. To further investigate this, we evaluate the decoding steps of {\method} across subsets of the MATH-500 benchmark categorized by difficulty levels. As shown in Fig.~\ref{fig:difficulty_level}, {\method} achieves progressively greater acceleration as the difficulty decreases, highlighting its capability to adaptively optimize inference speed based on task complexity.

\subsection{Analysis and Ablation Studies of WINO+}

\subsubsection{WINO+ improves base models under the same decoding budget}
To isolate the effect of trajectory injection on the model itself, we further evaluate the trained {\method}+ model under the standard decoding strategy, using the same 256 decoding steps as the base model. 
As shown in Table~\ref{tab:default_decoding_ablation}, {\method}+ consistently outperforms the corresponding base model under identical decoding budgets. 
On GSM8K, {\method}+ improves the accuracy from 73.24\% to 79.23\%, and on MMMU-val, it improves the accuracy from 18.56\% to 28.44\%. 
Since both models use the same standard decoding procedure and the same number of decoding steps, these gains cannot be attributed to a different inference algorithm or a larger decoding budget. 
Instead, the results indicate that WINO-guided trajectory injection improves the denoising behavior of the model itself, leading to better generation quality.

\subsubsection{On injected trajectories}
We compare different trajectory choices for trajectory injection in {\method}+ and summarize the results in Table~\ref{tab:trajectory_ablation}. The random-trajectory variant constructs intermediate supervision from randomly selected token-revealing orders, while the WINO-trajectory variant uses the verification-guided revealing steps produced by {\method}. As shown in the table, the quality of the injected trajectory plays a critical role. On GSM8K, replacing random trajectories with WINO trajectories improves the accuracy from 72.63\% to 76.58\%, while reducing the average decoding steps from 46.69 to 37.47. A similar trend is observed on MMMU-val, where the accuracy increases from 26.67\% to 28.11\% and the average decoding steps decrease from 45.90 to 26.25. These results suggest that the improvement of {\method}+ is not simply due to adding intermediate partially masked states during training. Instead, the trajectory produced by {\method} carries useful order information from its draft--verify--fallback process. By using such trajectories for training, {\method}+ provides a more suitable denoising order than random reconstruction, making the training process better aligned with efficient inference.
\begin{table}[t!]
\centering
\caption{Accuracy comparison between the base model and the trained {\method}+ model under standard decoding.}
\label{tab:default_decoding_ablation}
\renewcommand{\arraystretch}{1.25}
\resizebox{0.98\columnwidth}{!}{
\begin{tabular}{@{}!{\vrule width 0pt}p{2.1cm}<{\centering}p{1.8cm}<{\centering}p{1.5cm}<{\centering}p{1.2cm}<{\centering}}
\toprule
\textbf{Benchmark} & \textbf{Model} & \textbf{Accuracy} & \textbf{Steps} \\
\midrule
\multirow{2}{*}{GSM8K}    & LLaDA    & 73.24 & 256 \\
                          & WINO+    & 79.23 & 256 \\
\midrule
\multirow{2}{*}{MMMU-val} & MMaDA    & 18.56 & 256 \\
                          & WINO+    & 28.44 & 256 \\
\bottomrule
\end{tabular}
}
\vspace{-5pt}
\renewcommand{\arraystretch}{1.0}
\end{table}

\begin{table*}[t!]
\centering
\caption{Ablation study of trajectories.}
\label{tab:trajectory_ablation}
\renewcommand{\arraystretch}{1.25}
\resizebox{0.98\textwidth}{!}{
\begin{tabular}{@{}!{\vrule width 0pt}p{2.3cm}<{\centering}p{2.3cm}<{\centering}p{1.5cm}<{\centering}p{1.3cm}<{\centering}p{1.8cm}<{\centering}p{1.3cm}<{\centering}p{1.6cm}<{\centering}}
\toprule
\multirow{2}{*}{\textbf{Benchmark}} & \multirow{2}{*}{\textbf{Trajectory}} & \multirow{2}{*}{\textbf{Accuracy}} & \multirow{2}{*}{\textbf{Steps}} & {\textbf{Step}} & \multirow{2}{*}{\textbf{TPS}} & {\textbf{TPS}} \\
& & & & \textbf{Reduction} & & \textbf{Speedup} \\
\midrule
\multirow{2}{*}{GSM8K} & Random trajectory & 72.63 & 46.69 & 5.48 $\times$ & 96.62 & 5.44 $\times$ \\
& WINO trajectory & 76.58 & 37.47 & 6.83 $\times$ & 121.86 & 6.86 $\times$ \\
\midrule
\multirow{2}{*}{MMMU-val} & Random trajectory & 26.67 & 45.90 & 5.58 $\times$ & 31.27 & 5.19 $\times$ \\
& WINO trajectory & 28.11 & 26.25 & 9.75 $\times$ & 54.18 & 9.00 $\times$ \\
\bottomrule
\end{tabular}
}\vspace{-5pt}
\renewcommand{\arraystretch}{1.0}
\end{table*}

\begin{table*}[t!]
\centering
\caption{Training objective ablation of {\method}+ with different loss components.}
\label{tab:training_method_ablation}
\renewcommand{\arraystretch}{1.45}
\resizebox{\textwidth}{!}{
\begin{tabular}{@{}!{\vrule width 0pt}p{1.4cm}<{\centering}p{1.7cm}<{\centering}p{1.7cm}<{\centering}p{1.7cm}<{\centering}p{1.7cm}<{\centering}p{1.4cm}<{\centering}p{1.3cm}<{\centering}p{1.8cm}<{\centering}p{1.3cm}<{\centering}p{1.6cm}<{\centering}}
\toprule
\multirow{2}{*}{\textbf{Method}} & \multirow{2}{*}{\textbf{Benchmark}} & \multicolumn{3}{c}{\textbf{Training Loss}} & \multirow{2}{*}{\textbf{Accuracy}} & \multirow{2}{*}{\textbf{Steps}} & {\textbf{Step}} & \multirow{2}{*}{\textbf{TPS}} & {\textbf{TPS}} \\
& & $\mathcal{L}_{\mathrm{tok}}$ & $\mathcal{L}_{\mathrm{defer}}$ & $\mathcal{L}_{\mathrm{sharp}}$ & & & \textbf{Reduction} & & \textbf{Speedup} \\
\midrule
\multirow{4}{*}{LLaDA} & \multirow{4}{*}{GSM8K} & $\checkmark$ &  &  & 73.16 & 42.28 & 6.05 $\times$ & 109.05 & 6.14 $\times$ \\
& & $\checkmark$ & $\checkmark$ &  & 75.59 & 39.60 & 6.46 $\times$ & 115.62 & 6.51 $\times$ \\
& & $\checkmark$ &  & $\checkmark$ & 72.71 & 40.14 & 6.38 $\times$ & 114.91 & 6.47 $\times$ \\
& & \cellcolor{colorours}$\checkmark$ & \cellcolor{colorours}$\checkmark$ & \cellcolor{colorours}$\checkmark$ & \cellcolor{colorours}76.58 & \cellcolor{colorours}37.47 & \cellcolor{colorours}6.83 $\times$ & \cellcolor{colorours}121.86 & \cellcolor{colorours}6.86 $\times$ \\
\midrule
\multirow{4}{*}{MMaDA} & \multirow{4}{*}{MMMU-val} & $\checkmark$ &  &  & 25.56 & 44.98 & 5.69 $\times$ & 31.99 & 5.31 $\times$ \\
& & $\checkmark$ & $\checkmark$ &  & 26.00 & 33.70 & 7.60 $\times$ & 42.41 & 7.06 $\times$ \\
& & $\checkmark$ &  & $\checkmark$ & 25.22 & 35.63 & 7.18 $\times$ & 41.29 & 6.86 $\times$ \\
& & \cellcolor{colorours}$\checkmark$ & \cellcolor{colorours}$\checkmark$ & \cellcolor{colorours}$\checkmark$ & \cellcolor{colorours}28.11 & \cellcolor{colorours}26.25 & \cellcolor{colorours}9.75 $\times$ & \cellcolor{colorours}54.18 & \cellcolor{colorours}9.00 $\times$ \\
\bottomrule
\end{tabular}
}
\renewcommand{\arraystretch}{1.0}
\end{table*}

\subsubsection{On WINO+ training objectives}
We study the contribution of each training objective in {\method}+ and report the results in Table~\ref{tab:training_method_ablation}. Using only $\mathcal{L}_{\mathrm{tok}}$ already brings clear acceleration, showing that the WINO-derived revealing order provides useful supervision. However, it only specifies which tokens to decode at each step, without constraining positions that should remain masked. Adding $\mathcal{L}_{\mathrm{defer}}$ suppresses high-confidence wrong predictions on deferred positions, improving the accuracy from 73.16\% to 75.59\% on GSM8K and from 25.56\% to 26.00\% on MMMU-val. This indicates that learning which tokens should not be revealed is also important. Finally, when combined with $\mathcal{L}_{\mathrm{tok}}$ and $\mathcal{L}_{\mathrm{defer}}$, $\mathcal{L}_{\mathrm{sharp}}$ further refines the confidence of reliable predictions. The full objective achieves the best trade-off, reaching 76.58\% with 37.47 steps on GSM8K and 28.11\% with 26.25 steps on MMMU-val.

\begin{figure}[t]
    \centering
    \includegraphics[width=\columnwidth]{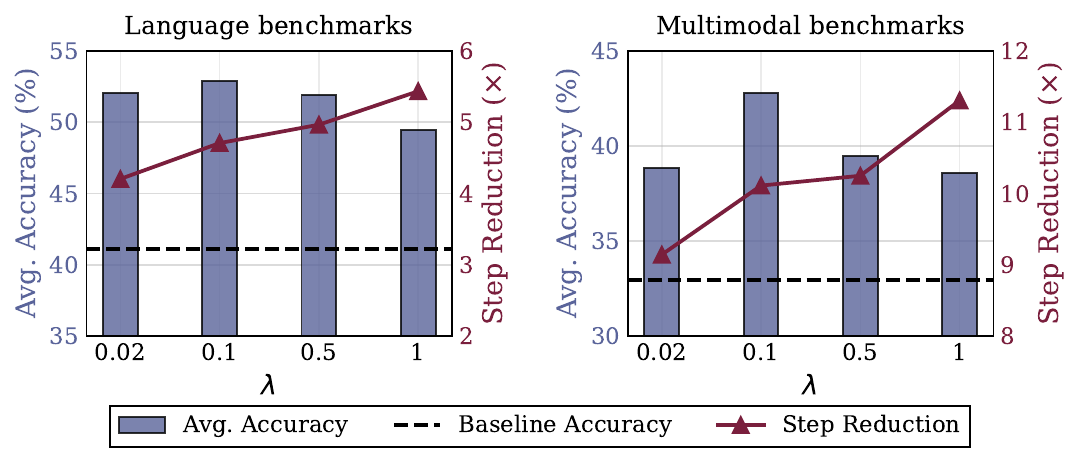}
    \caption{Ablation of the loss-balancing coefficient $\lambda$ in {\method}+. The plots report average accuracy and decoding-step reduction across language and multimodal benchmarks under different $\lambda$ values.}
    \label{fig:ablation_sharp_weight}
\end{figure}

\subsubsection{On loss weight}
We study the effect of the sharpening loss by varying $\lambda$ over $\{0.02, 0.1, 0.5, 1.0\}$.
Fig.~\ref{fig:ablation_sharp_weight} reports the averaged accuracy and decoding-step reduction over language benchmarks and multimodal benchmarks. As shown in the figure, $\lambda=0.1$ achieves the best accuracy on both language and multimodal benchmarks, suggesting that a moderate sharpening signal helps the model confidently reveal stable tokens early. 
Increasing the weight further improves step reduction, but hurts accuracy by encouraging overly aggressive parallel revelation. 
Thus, we use $\lambda=0.1$ as the default setting, which offers the best trade-off between quality and efficiency.

\subsection{Efficiency and Case Study}
\subsubsection{GPU memory usage}
\begin{figure}[t!]
    \centering
    \includegraphics[width=0.78\columnwidth]{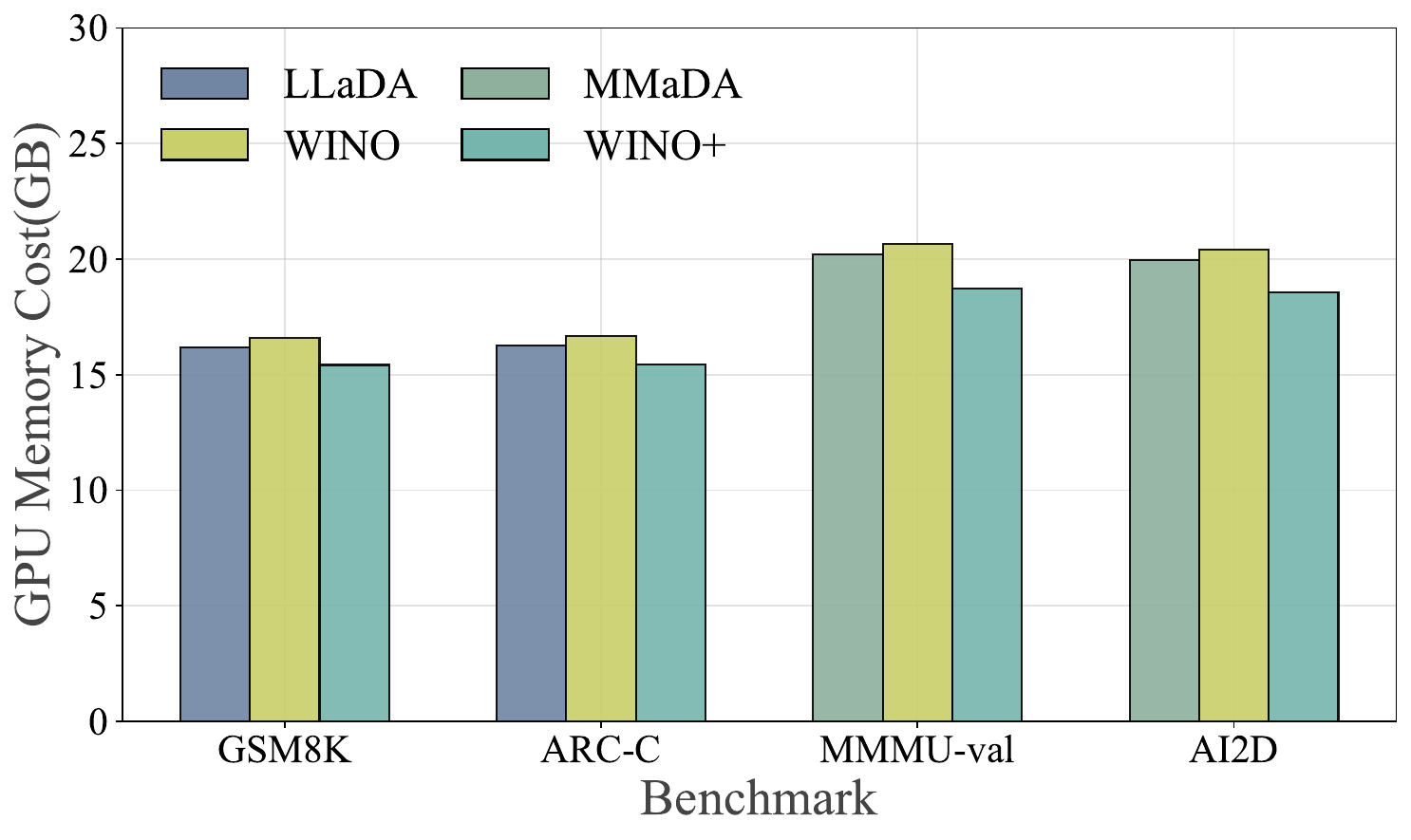}
    \vspace{-8pt}
    \caption{Peak GPU memory usage of the baseline DLLMs, {\method}, and {\method}+ across representative language and multimodal benchmarks.}
    \label{fig:gpu_mem}
\end{figure}

We compare the peak GPU memory usage of the standard baselines, {\method}, and {\method}+ in Fig.~\ref{fig:gpu_mem}. To facilitate efficient verification of unmasked tokens, {\method} introduces an auxiliary shadow block whose size equals the specified block length $L_b$ of the semi-autoregressive decoding process. Therefore, {\method} induces additional GPU memory cost due to the longer effective sequence length. Nevertheless, this overhead remains marginal compared to the baselines across the benchmarks. For instance, on GSM8K, {\method} increases GPU memory usage by only 2.4\% compared to standard LLaDA decoding, from 16.18 GiB to 16.57 GiB. In contrast, {\method}+ removes the shadow block at inference time by injecting the stabilized-prediction behavior into the model parameters, reducing the memory usage to 15.42 GiB on GSM8K, lower than both {\method} and LLaDA. Across both language and multimodal benchmarks, {\method}+ consistently achieves the lowest memory footprint. This indicates that {\method}+ converts the online verification behavior of {\method} into a more memory-efficient inference procedure, rather than trading additional memory for speed.

\begin{figure*}[!t]
    \centering
    \includegraphics[width=\textwidth]{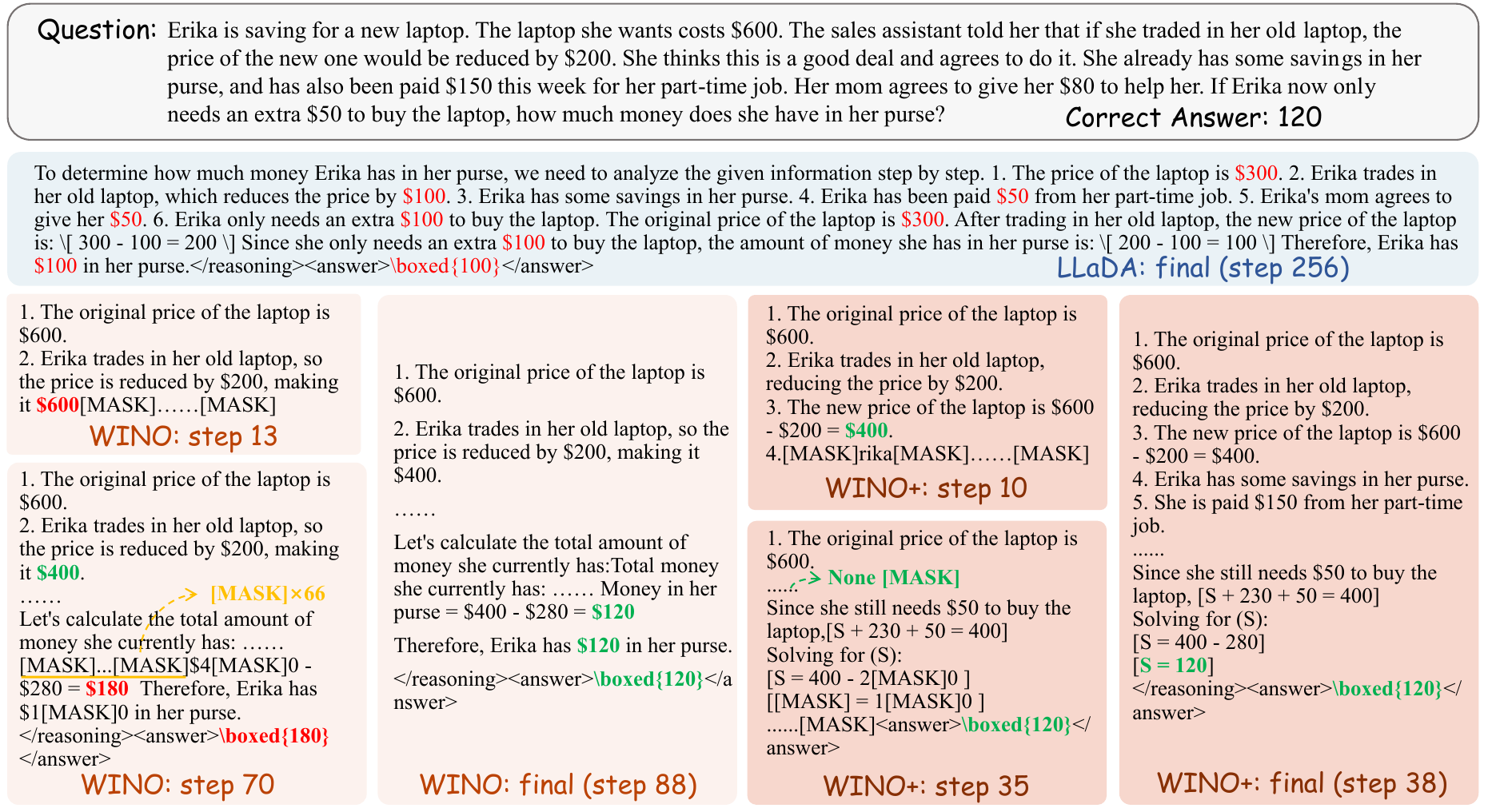}
    \caption{Decoding trace on a GSM8K example for standard decoding, {\method}, and {\method}+. The figure shows selected intermediate and final outputs at different decoding steps. Red text indicates incorrect intermediate or final generated results, green text indicates the correct final result, and [MASK] denotes positions that remain undecoded.}
    \label{fig:case-gsm8k-wino-plus}
\end{figure*}

\subsubsection{Case Study}
To conduct a fine-grained examination of decoding dynamics, we present a GSM8K example in Fig.~\ref{fig:case-gsm8k-wino-plus}. 
The standard decoding baseline may produce erroneous tokens at early decoding stages. 
Since these generated tokens remain unchanged in subsequent decoding steps, the false contextual information can propagate throughout the generation process and eventually lead to low-quality outputs. 
In contrast, {\method} enables dynamic refinement through its iterative draft-and-verify mechanism: unstable predictions can be identified, remasked, and regenerated when richer context becomes available, thereby mitigating error accumulation.

The behavior of {\method}+ further shows that such dynamic refinement can be transferred into the model through trajectory injection. 
Rather than using shadow block for online verification, {\method}+ learns from the trajectories produced by {\method} and forms a more suitable denoising order during training.
As shown in the example, {\method}+ keeps uncertain reasoning-critical positions masked until sufficient context is available, while generating stable tokens earlier. 
As a result, it reaches the correct answer with substantially fewer decoding steps than both the standard baseline and {\method}. 
This case illustrates how WINO-derived trajectories help {\method}+ preserve the refinement benefit of {\method} while enabling a simpler and faster inference process.

\section{Conclusion}

In this paper, we studied the quality and efficiency trade-off in open-source Diffusion Large Language Models and attributed it to a train-inference mismatch amplified by irreversible decoding. While standard training reconstructs tokens from randomly corrupted states, efficient inference requires an adaptive denoising order that commits reliable tokens early and defers context-dependent ones. To address this issue, we proposed WINO, a training-free revokable decoding algorithm with a Wide-In, Narrow-Out draft, verify, and fallback mechanism. WINO enables aggressive parallel drafting while re-evaluating generated tokens under enriched bidirectional context and re-masking unreliable ones, thereby improving both generation quality and decoding efficiency. We further introduced WINO+, a trajectory-injection framework that distills the verified finalization order discovered by WINO into model parameters. By replacing random reconstruction with trajectory-ordered denoising, WINO+ teaches the model when each token should be revealed or deferred and reduces the need for online rollback. Experiments on LLaDA and MMaDA across language and vision-language benchmarks show that WINO consistently improves parallel decoding, while WINO+ further strengthens the quality-efficiency trade-off by internalizing verified denoising trajectories. These results suggest that DLLMs can serve as their own efficiency teachers, using revokable decoding to discover reliable generation orders and trajectory injection to learn them for faster, higher-quality generation.

\bibliographystyle{IEEEtran}
\bibliography{main}
\newpage
\appendices

\section{Standard DLLM Training Objective}
\label{app:std-dllm-training}

Given a prompt $X$ and a reference response 
$Y^\star=[y_1^\star,\ldots,y_L^\star]$, standard DLLM training samples a masking ratio 
$\rho\sim\mathcal{U}(0,1)$ and independently masks each response token with probability $\rho$, while keeping the prompt unchanged. 
Let $\widetilde{Y}^{(\rho)}$ denote the corrupted response, and let 
$M_\rho=\{l\in [L]\mid \widetilde{y}^{(\rho)}_l=\texttt{[MASK]}\}$ be the set of masked response positions, where $[L]=\{1,\ldots,L\}$.
Following the standard masked diffusion formulation used in LLaDA and MMaDA~\cite{DBLP:conf/nips/ShiHWDT24,DBLP:conf/nips/SahooASGMCRK24,DBLP:journals/corr/abs-2502-09992,DBLP:journals/corr/abs-2505-15809}, the objective is
\begin{equation}
\begin{aligned}
\mathcal{L}_{\mathrm{std}}(\theta)
=
-\mathbb{E}
\left[
\frac{1}{\rho}
\sum_{l\in M_\rho}
\log p_\theta
\left(
y_l^\star
\mid X,\widetilde{Y}^{(\rho)}
\right)
\right],
\end{aligned}
\end{equation}
where the expectation is taken over 
$(X,Y^\star)\sim\mathcal{D}$, 
$\rho\sim\mathcal{U}(0,1)$, and 
$\widetilde{Y}^{(\rho)}$ sampled from the corresponding masking process.
Here, $Y^\star$ denotes the reference response from the training data.
This loss supervises what token should fill each masked position, but not when the token should be revealed.

\section{More Training Details for WINO+}
\label{app:wino_plus_training}

We provide additional training details for WINO+. All WINO+ variants are trained with parameter-efficient LoRA fine-tuning~\cite{DBLP:conf/iclr/HuSWALWWC22} on WINO-guided trajectories using bf16 mixed precision. For LLaDA-8B-Instruct, we collect WINO trajectories from GSM8K~\cite{DBLP:journals/corr/abs-2110-14168} and Countdown-Tasks-3to4~\cite{DBLP:journals/corr/abs-2504-12216}, retaining only samples with correct final answers. This yields about 5.6K GSM8K pairs and 0.4K Countdown pairs before step-wise expansion. We train the model sequentially, first on the easier GSM8K trajectories and then on the more challenging Countdown trajectories. For MMaDA-8B-MixCoT, we construct trajectories from a subset of the IconQA~\cite{DBLP:conf/nips/LuQCXZZYLZ21} training set and obtain about 0.7K question-answer pairs before expansion. All trajectory data are constructed from training-set samples only.

For each retained sample, the final WINO response is used as the target, and all WINO steps are expanded as ordered training instances. Offline trajectories use block size 128, $\tau_1=0.6$, and $\tau_2=0.9$. WINO+ does not use the shadow-block verifier at inference.

\begin{table}[htbp]
\centering
\caption{Optimization configuration for WINO+ training.}
\label{tab:wino_plus_optim}
\begin{tabular}{lcc}
\toprule
Configuration & LLaDA-8B-Instruct & MMaDA-8B-MixCoT \\
\midrule
Precision & bf16 & bf16 \\
Optimizer & AdamW & AdamW \\
Learning rate & $2\times10^{-5}$ & $1\times10^{-4}$ \\
Weight decay & 0.01 & 0.05 \\
LR scheduler & constant & cosine \\
Training length & 6 epochs & 1500 steps \\
Batch size & 1 & 1 \\
Gradient accumulation & 8 & 16 \\
Max grad norm & 1.0 & 1.0 \\
Warmup steps & -- & 500 \\
Minimum LR scale & -- & 0.1 \\
\bottomrule
\end{tabular}
\end{table}

LoRA adapters are inserted into \texttt{q\_proj}, \texttt{k\_proj}, \texttt{v\_proj}, \texttt{o\_proj}, \texttt{gate\_proj}, \texttt{up\_proj}, and \texttt{down\_proj}. The LoRA rank and scaling factor are both set to 128, and the bias term is not adapted. We use a LoRA dropout of 0.0 for LLaDA and 0.05 for MMaDA. The balance weight $\lambda$ in the WINO+ training objective is set to 0.1 for all experiments. The optimization configurations are summarized in Table~\ref{tab:wino_plus_optim}.

All WINO+ training experiments are conducted on 8 NVIDIA RTX 3090 GPUs with DeepSpeed-based distributed optimization~\cite{rajbhandari2020zero}. We use CPU offloading and communication-overlap optimization to reduce memory overhead during LoRA fine-tuning.
\vfill

\end{document}